FRONT MATTER

**Title**

A Shank Angle-Based Control System Enables Soft Exoskeleton to Assist Human Non-Steady Locomotion

**Authors**

Xiaowei Tan[1], Weizhong Jiang[1,2], Bi Zhang[1], Wanxin Chen[1,2], Yiwen Zhao[1], Ning Li[1]*, Lianqing Liu[1], Xingang Zhao[1]*

**Affiliations**

[1]State Key Laboratory of Robotics and Intelligent Systems, Shenyang Institute of Automation, Chinese Academy of Sciences, Shenyang 110016, China.
[2]University of Chinese Academy of Sciences, Beijing 100049, China.

* Address correspondence to: lining3@sia.cn (N.L.); zhaoxingang@sia.cn (X.Z.)

**Abstract**

Exoskeletons have been shown to effectively assist humans during steady locomotion. However, their effects on non-steady locomotion, characterized by nonlinear phase progression within a gait cycle, remain insufficiently explored, particularly across diverse activities. This work presents a shank angle-based control system that enables the exoskeleton to maintain real-time coordination with human gait, even under phase perturbations, while dynamically shaping assistance profiles to match the biological ankle moment patterns across walking, running, stair negotiation tasks. The control system consists of an assistance profile online generation method and a model-based feedforward control method. The assistance profile is formulated as a dual-Gaussian model with the shank angle as the independent variable. Leveraging only IMU measurements, the model parameters are updated online each stride to adapt to inter- and intra-individual biomechanical variability. The profile tracking control employs a human-exoskeleton kinematics and stiffness model as a feedforward component, reducing reliance on historical control data due to the lack of clear and consistent periodicity in non-steady locomotion. Three experiments were conducted using a lightweight soft exoskeleton with multiple subjects. The results validated the effectiveness of each individual method, demonstrated the robustness of the control system against gait perturbations across various activities, and revealed positive biomechanical and physiological responses of human users to the exoskeleton's mechanical assistance.

**MAIN TEXT**

**INTRODUCTION**

Robotic exoskeletons serve various purposes, from augmenting physical abilities of workers [1, 2], soldiers [3, 4], and normal people [5, 6], to restoring mobility in patients with motor dysfunction [7, 8] and aiding daily activities [9, 10]. Unlike autonomous robots, exoskeletons collaborate physically and functionally with users towards a common task, which raises further requirements for adaptability to various users [11], locomotion patterns [12], and natural behaviors [13]. Low adaptabilities can lead to counterproductive effects. Significant research has focused on improving exoskeleton adaptability from the aspects of structure and control system.

Exoskeletons' structure typically contains rigid, heavy materials and mechanical joints that are desired to align with human joints and provide rotational assistance. Any motion mismatches will be perceived by the user due to high interface stiffness [14], making them suitable mainly for individuals with low proprioception, such as paraplegic patients [15], or task-centered workers [2]. There are studies that integrated soft or elastic components, such as fabrics [16, 17], cables [18, 19], or elastic actuators [20, 21], into the structure to improve safety and functionality. Soft exoskeletons do not rely on mechanical joints; instead, they assist movement by synchronizing cable- or pneumatic-based linear actuator motions with human muscle contraction and relaxation, providing more natural, biomimetic assistance to reduce biological muscle effort [12, 22-24]. Although, the inherent low stiffness of soft materials helps mitigate the discomfort induced by the motion mismatches, soft exoskeletons still require the ability to actively adapt to changes in users and their activities.

Robot control systems are key to the active adaptability of exoskeletons. The procedures for exoskeleton control system implementation [10, 25, 26] typically consist of three stages: decoding human or environment information from sensor signals, such as kinematic or bioelectrical signals, adjusting exoskeleton behavior profiles based on the decoding results, and finally controlling the exoskeleton actuators to produce the exact profiles. However, the cascaded architecture linking the three stages often requires pre-pairing among them, resulting in reduced robustness and increased system complexity, especially when involving diverse human activities and terrains. Thus, a unified and robust control strategy capable of coping with various human activities without relying on frequent internal mode switching is desired for exoskeletons. Moreover, exoskeletons' behavior profile is crucial, as it determines the final assistive forces delivered to the human body, provided the controller can accurately track the profile. The profile is determined by its model and independent or progression variable. An ideal profile should be based on a model that accurately reflects human biomechanics, while ensuring the continuous synchronization of its independent variable (e.g., time or percentage of the gait cycle [GC], as commonly used) with human gait progression, particularly during non-steady locomotion. Non-steady locomotion refers to the activities where progression changes nonlinearly and irregularly, such as acceleration, deceleration, recovery from external disturbances, and negotiating variations in terrain [27], [28]. A control system that integrates a unified architecture, a biomimetic profile model, and a highly synchronized independent variable has strong potential to advance exoskeleton applications in complex daily activities and environments.

Some researchers explored designing exoskeleton behavior profiles directly and instantly from data such as plantar pressure [29, 30], surface electromyography (sEMG) [31, 32], metabolic rate [33, 34], or muscle ultrasound images [35, 36], rather than switching profiles correspondingly to each classified activity type. The human-in-the-loop strategy in the exoskeleton field, proposed by Koller [37] and Zhang [34], offers an effective way to eliminate the intention recognition layer from the control architecture. Instead of switching exoskeleton behavior profile models based on classified activity types, it adjusts the models according to the user's needs, which are estimated from physiological signals such as metabolic rates [34] or sEMG [32]. However, these works focused on the profile models and paid less attention to the role of profile independent variables, often using time or percent GC to control the progression or movement of force point along the profile. These variables have a constant rate of change within one GC, i.e., 1 and $1/T$, where $T$ is the gait period. Time is irreversible, independent, and unaffected by any external factors. The variables are not suitable for human non-steady locomotion, where human gait progression changes nonlinearly even within one GC. For example, the gait toe off event may occur at 60% or 20% of the GC, depending on the specific non-steady situation. Consequently, several recent studies focused on finding or designing variables that can change continuously in coordination with the progression of human non-steady gait [38-40]. Villarreal [39] explored the potential of using hip phase angle for gait parameterization under disturbances. Macaluso [40] compared the effectiveness of thigh phase angle, tibia phase angle, and time in quantifying human gait progression during non-steady slope walking. Tan [41] improved the calculation of thigh phase angle to enhance its robustness across different individuals and environments. However, these phase angles were proposed for describing human gait, without in-depth consideration of the exoskeleton behavior profiles' models and design of control systems. As a result, despite recent advances, most exoskeleton control systems still fall short of meeting all the key requirements: adapting behavior profile models to human biomechanical needs without pre-classifying activity types, and ensuring that independent variables remain synchronized with human gait progression, even during non-steady locomotion across diverse activities.

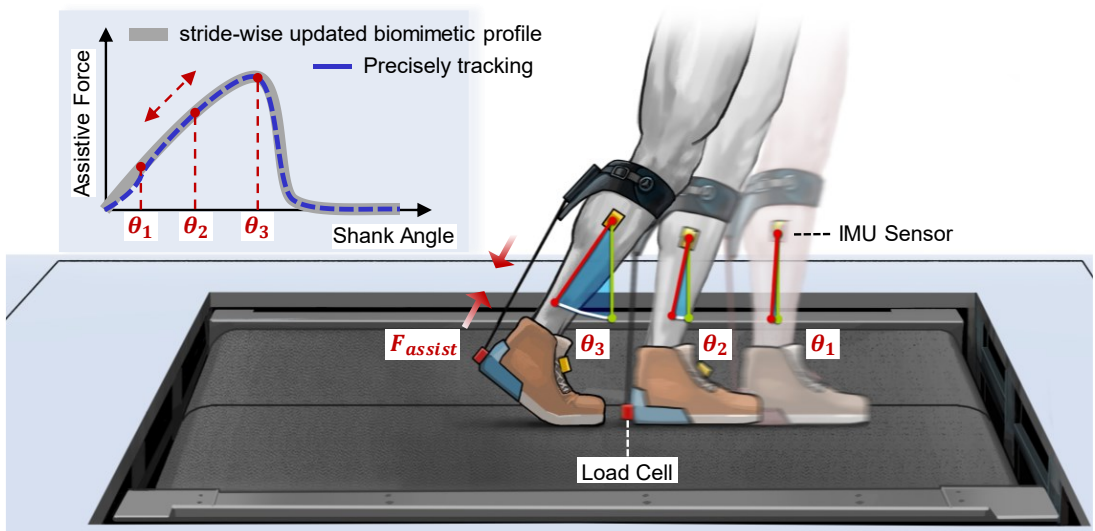

**Fig. 1.** Overall diagram of the shank angle-based exoskeleton control system. Human shank angle ($\theta$) is used to control the movement of force point ($F_{assist}$) along the assistance profile. The dual-Gaussian-based assistance profile is stride-wise updated using only IMU measurements to ensure consistent alignment with the user's biological ankle torque patterns across various types of activities. The exoskeleton, utilizing the designed control system, adjusts the length of the actuation cable parallel to the Achilles tendon to precisely produce the profiles.

In this work, we present a time-independent, versatile control system for controlling a soft ankle exoskeleton to provide biomimetic ankle plantar flexion (PF) assistance during the stance phase (Fig. 1), even under complex non-steady conditions across walking, running, and stair negotiation activities, without the need for explicit classification of activity types. The soft exoskeleton is developed with lightweight, flexible, and durable leather materials and Bowden cable-driven actuators. It assists user movement by providing assistive forces parallel to the Achilles tendon and generating assistive torques based on the user's own ankle bone joint. In the design of the control system, rather than using time, we propose using human shank angle, i.e., the shank to vertical angle, as a variable to parameterize the progression of assistance profiles, which enables consistent pairwise alignment between the desired assistive forces and the human gait states, even under non-steady locomotion conditions. A dual-Gaussian-based profile model is designed to reflect the pattern of human biological ankle PF torques. Leveraging the analysis of human gait biomechanics and mechanical energy transformation, it is found that the peak human ankle dorsiflexion (DF) angle consistently coincides with the peak ankle PF torque, regardless of terrains (level ground and slope) or activities (walking and running). We leverage this finding and design a method to update the model parameters online every GC using only IMU measurements, which enhances the instant adaptabilities of the control system to variations in human biomechanics without requiring advanced instruments. To ensure precise tracking of the designed profile, a model-based feedforward control method is developed that does not rely on any historical control information, making it ideal for non-steady activities where clear rhythmicity is absent.

We hypothesize that the shank angle-based control system can generate assistance profiles consistently synchronized with human gait states and biomechanics, even during non-steady conditions and diverse types of activities, and improve exoskeletons' performance in assisting human movement. Experiments consisting of three sessions were conducted to test the hypothesis and verify the profile generation and tracking abilities of the proposed control system, the synchronization and biomimetic properties of the profile under non-steady conditions, and the positive effects of the exoskeleton assistance on human physiology and biomechanics during level walking (LW), level running (LR), ramp ascent (RA), and ramp descent (RD) activities.

## MATERIALS AND METHODS
### A Soft Ankle Exoskeleton Platform Design

Our laboratory previously developed portable [12] and non-portable [22] soft exoskeletons using lightweight, flexible, and durable leather materials and Bowden cable-driven actuators. Here, we present a new soft ankle exoskeleton that leaves the user's anterior and posterior shank muscles exposed to facilitate sEMG electrode placement. It incorporates a BOA lacing system and magnetic buckles, reducing donning and doffing times to 24 s and 10 s, respectively (see Supplementary Movie S1). A bill of materials for the overall platform is provided in Supplementary Data File S1.

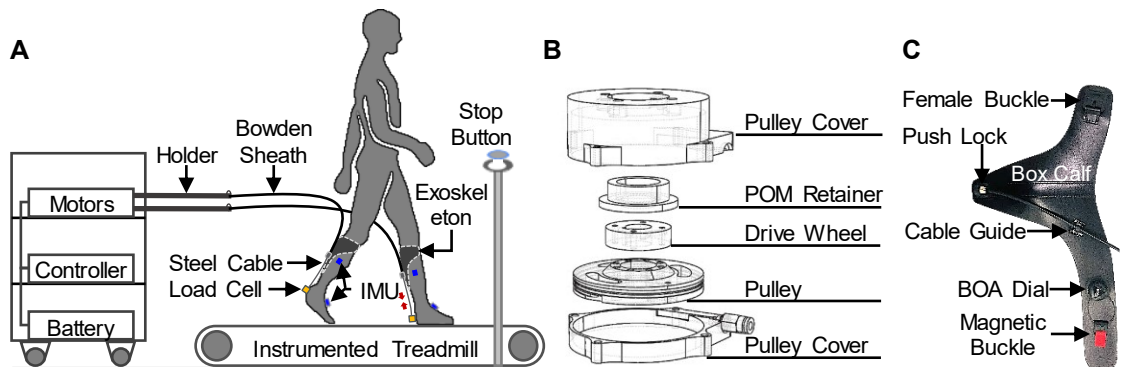

**Fig. 2.** Overview of the soft exoskeleton platform. (A) Soft ankle exoskeleton platform. It applied the proposed control system for experiments. It consists of wearable lightweight suit with integrated sensors and off-board actuation system positioned behind the exoskeleton user. Actuation steel cables, housed in a protective sheath, transmit mechanical force to the user. (B) Customized cable pulley cartridge. Gearbox outer case is mounted to the pully cover and its shaft connects to the pulley wheel through a drive wheel. A retainer made from low-friction polyoxymethylene (POM) material is positioned between the cover and the drive wheel to minimize possible friction. (C) Right-side clothing suit. It is lightweight, flexible, and easy to don and doff. The suit is crafted from box calf leather and integrates a BOA lacing system and a pair of magnetic buckles for rapid adjustment.

As presented in Fig. 2A, the soft exoskeleton applies an off-board actuation system placed behind the treadmill and user to isolate the effect of additional loads on human gait characteristics. This actuation system consists of an embedded controller (PCM-3365, Advantech, China), a 24 V battery (11.6 Ah, Jin Feng, China), and two sets of gear motors (EC-4pole with 66:1 reduction ratio, Maxon, USA) paired with drivers (Gold Solo Twitter, Elmo Motion Control, Israel). A customized cable pulley cartridge (Fig. 2B) is fixed to the output side of each gear motor. The pulley component is coupled with the gear motor's shaft to convert motor torques into linear forces transmitted through a steel cable. The cartridge is specifically designed to prevent cable derailment from the intended pulley slot. The controller operates RT Ubuntu 14.04 and sends velocity commands to the motor drivers via an EtherCAT fieldbus at a rate of 1kHz. One end of the Bowden cable sheath is secured to the pulley cover using a push-to-connect fitting (PC5-M5C, Zhuo Ji Qi Dong, China) and supported by a holder to minimize cable bending and reduce friction and energy loss.

Another end of the sheath is anchored to the clothing suit (Fig. 2C). The steel cable comes out from the sheath and connects to a miniature load cell that is fixed to the shoe heel counter. The length of the steel cable between the two anchor points will shorten when the motor rotates to wind the cable around the pulley, and forces are generated on the cable to assist user ankle PF muscles. The clothing suit is primarily made of box calf leather and is 0.067 kg per leg in mass and supports users with height of 1.6 m - 1.9 m. The BOA lacing system and magnetic buckles allow quick fastening and easy adjustment of the suit. The push-to-connect lock and cable guide clamp are used to secure and orient the cable sheath, respectively. The suit structure is specifically designed to minimize compression on the user's shank muscles, which is crucial for obtaining stable muscle activation signals and accurately evaluating exoskeleton performance.

A single load cell (LSB205, Futek, USA) and two wireless inertial measurement unit (IMU) sensors (Xsens Awinda, Xsens, The Netherlands) are deployed on the user's each leg to measure the delivered mechanical forces and human motions, which are necessary for a closed loop control system. The load cell signals are processed by a digital amplifier (USB220, Futek, USA) and read by the controller at a rate of 1 kHz. The two IMU sensors are positioned on the user's lateral shank and shoe instep respectively, transmitting data wirelessly to the controller at a rate of 100 Hz. Additionally, a split-belt treadmill (Fully Instrumented Treadmill, Bertec, USA) is utilized in the platform, allowing independent control of each belt's motion parameters, such as velocity, acceleration, and direction. This function will facilitate simulation of non-steady conditions.

**Shank Angle-Based Assistance Profile Design**
The shank angle, i.e., the shank to vertical angle, is a great candidate variable for time to represent the progression of the human gait stance period [22]. Please note that the analyses of human ankle kinematics and kinetics presented in this section are conducted based on a publicly available multi-subject dataset [42]. As shown in Fig. 3A, in this study, we design assistance profile model that applies shank angle as the independent variable and utilizes two Gaussian curves to describe the profile's rising and falling parts. The Gaussian model was chosen for its alignment with the single-crest nature of human ankle torques, and for the ease with which its shape can be directly controlled through a few key parameters. The two curves are supposed to imitate the rising and falling parts of the biological ankle PF torques respectively, with the curves joining at the peak PF torque point. In this study, shank angle values are defined as zero when the human body is in an upright posture and positive when the knee is flexed. Thus, shank angle transitions from negative to positive as the gait stance period progresses from foot contact to foot-off. The proposed exoskeleton force profile is parameterized as follows:

$$F(\theta) = \begin{cases} A \cdot \exp\left(-\dfrac{(\theta - \mu)^2}{2\sigma_1^2}\right) & \theta_{FC} < \theta \leq \mu \\[3mm] A \cdot \exp\left(-\dfrac{(\theta - \mu)^2}{2\sigma_2^2}\right) & \mu < \theta < \theta_{FO} \end{cases} \qquad (1)$$

where $F(\theta)$ is the desired assistive force at the shank angle of $\theta$, $A$ and $\mu$ are the amplitude and mean parameters of the two Gaussian curves, $\sigma_1$ and $\sigma_2$ are the standard deviation parameters of the first (rising) and second (falling) Gaussian curves, and $\theta_{FC}$ and $\theta_{FO}$ are the shank angles at the foot contact and foot-off events, respectively.

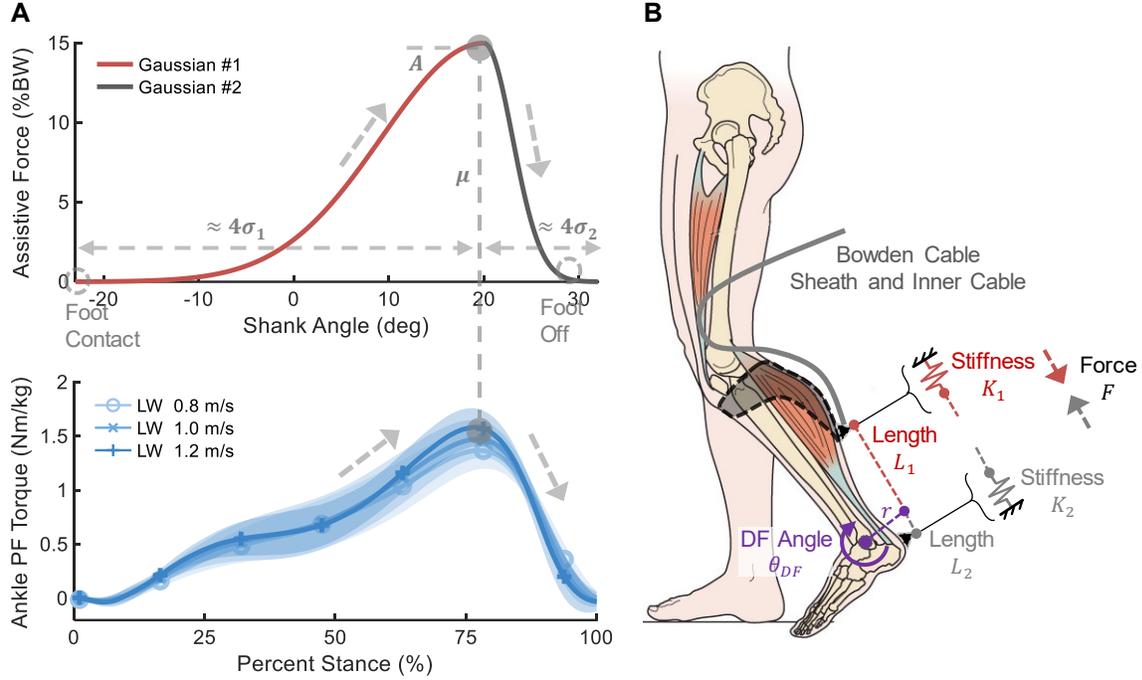

**Fig. 3.** Assistance profile and human-exoskeleton coupled modelling. (A) Designed assistance profile that utilizes two Gaussian curves joined at the peak point and applies shank angle as the independent variable. The profile exhibits high similarities to the human biological ankle PF torques. The shaded area on the PF torques denotes the standard deviation across subjects. (B) Human-exoskeleton coupled kinematics and stiffness modelling. The actuation cable between the end of the cable sheath and the end of the inner cable functions like an artificial Achilles tendon. The length of the artificial tendon is affected by the upper length ($L_1$), lower length ($L_2$), corresponding stiffness ($K_1$) and ($K_2$), forces on the inner cable ($F$), mechanical assistance lever arm ($r$), and ankle DF angle ($\theta_{DF}$).

**Profile Model Parameter Online Updating**

Human biological ankle torque patterns naturally vary with individuals, tasks, terrain, and activities. A crucial consideration in designing a versatile assistance profile is that the profile model parameters must be capable of online self-adjustment to adapt these variations. There are two key points in parameter estimation: 1) the new parameters should align with the user's gait biomechanics, and 2) the estimation procedures should be consistent across different activity types; otherwise, an activity classification method will be necessary, and a cascaded framework will still be applied.

As shown in Fig. 3A, the profile begins at the foot contact event (shank angle $\theta_{FC}$), reaches the maximum at the peak ankle PF torque (shank angle $\mu$), and ends at the foot-off event (shank angle $\theta_{FO}$). Biological PF torques cannot be calculated online, but here only the shank angle corresponding to the peak PF torque is required. As shown in Fig. S1, multi-subject gait analysis reveals that the peak ankle PF torque consistently coincides with the peak ankle DF angle, regardless of activity types (walking or running) or terrain (various inclines). This point can also be concluded from gait mechanical energy analysis. Ankle joint reverses its rotational direction and produces a large DF torque at the peak DF angle. This torque does positive work to release the potential energy stored in the ankle during the early stance phase and to increase the leg's kinetic energy for later transition from stance to swing. Therefore, humans require the most ankle motion assistance at this point to minimize the participation of human biological effort. Based on this analysis, the parameter $\mu$ is estimated as $\theta_{MDF}$, namely the shank angle corresponding to the peak ankle DF angle.

The designed profile begins and ends at the foot contact and foot-off events, respectively. Identifying features that can determine the two events across different activities would be important to build an activity-type-adaptive profile. We previously found two stable foot motion features for detection of LW foot contact (maximum foot pitch angle, namely forefoot-up) and foot-off (minimum foot pitch angular velocity) events [12]. It can be seen from Fig. S2 that this finding still applies to RA, RD, and LR activities. Ankle PF power reaches its maximum just before the foot-off event, and meanwhile, both ankle DF and foot pitch angular velocities reach their minimum values [43]. Foot pitch angle gradually increases during the swing period in case of foot drop and reaches the maximum near the foot contact event. The peak point seeking method reported in [12] is used here to achieve shank angles, $\theta_{FC}$ and $\theta_{FO}$, corresponding to foot contact and foot-off events, respectively. The remaining profile parameters, $\sigma_1$ and $\sigma_2$, are estimated as follows:

$$\sigma_1 = \frac{(\theta_{MDF} - \theta_{FC})}{4} \tag{2}$$

$$\sigma_2 = \frac{(\theta_{FO} - \theta_{MDF})}{4} \tag{3}$$

where the denominator of four is used to smooth the profile and prevent abrupt force changes at the ends of the profile. $A$ is fixed to a percent value of the user's body weight (BW). The assistance profile model, Eq. (1), updates stride-by-stride only based on IMU measurements. The position and movement of the assistive force along the profile are determined by the user's current shank angle and can change nonlinearly or even in reverse to remain in coordination with the user's gait progression.

**Human-Exoskeleton Coupled Kinematics and Stiffness Modelling**
Soft exoskeletons have limited bandwidth and typically utilize feedforward information to enhance profile tracking performance. Historical data on human gait states and profile control are commonly utilized to form the feedforward component. However, this solution is not suitable for non-steady locomotion, since non-steady events disrupt the rhythmicity of human gaits, making the historical data less effective for future control compensation. In this study, we propose a method modelling the human-exoskeleton coupled kinematics and stiffness, which is used as the feedforward component in control system.

As shown in Fig. 3B, the soft exoskeleton's actuation cable functions as an artificial tendon parallel to the user's Achilles tendon. The length of the artificial tendon consists of two parts $(L_1, L_2)$, divided by the line of the lever arm $(r)$ of the assistive forces. The stiffness coefficients during force generation and transmission at each anchor point of the artificial tendon are denoted as $K_1$ and $K_2$. The value of assistive forces $(F)$ is determined by the length and stiffness. The length of $L_1$ is calculated as follows:

$$L_1[t] = L_1[t_0] - \frac{F[t]}{K_1} - \Delta L_1[t] \tag{4}$$

where $t$ and $t_0$ denote the present and the initial times, and $\Delta L_1$ denotes the downward migration of the suit caused by repetitive stretching. The square brackets "[ ]" are used to indicate the value of the variable at a specific time, rather than implying that the variable is a function of time. The length of $L_2$ is calculated as follows:

$$L_2[t] = L_2[t_0] - \frac{F[t]}{K_2} + r \cdot \theta_{DF}(t) \tag{5}$$

where the value of the ankle DF angle ($\theta_{DF}$) is defined as zero when the user is in a straight standing posture. The overall length of the artificial tendon is calculated as follows:

$$L[t] = \overbrace{r \cdot \theta_{DF}(t)}^{A[t]} - \overbrace{\frac{F[t]}{K_{all}}}^{B[t]} + \overbrace{L_1[t_0] + L_2[t_0] - \Delta L_1[t]}^{C[t]} \tag{6}$$

where $K_{all}$ is $[K_1 K_2 / (K_1 + K_2)]$. To derive an analytical expression, each component of the artificial tendon length, namely $A[t]$, $B[t]$, and $C[t]$, is analyzed as follows.

$A[t]$ represents the component of the artificial tendon that changes with ankle rotation. The lever arm $r$ is measured manually for each individual. The ankle DF rotation angle $\theta_{DF}$ is obtained using the two IMUs attached to the user's shank and foot segments.

$B[t]$ represents the component related to suit deformation, resulting from the assistive forces passing though the soft suit. $F[t]$ is measured by the load cell placed at the end of the cable. $K_{all}$ is identified through experiments, in which a subject wears the exoskeleton and positions the tested leg in the walking push off pose. Assistive forces are gradually increased from 5 N to 180 N and then slowly decreased to 5 N, with forces and cable length data recorded over ten cycles. The data fitting results, as shown in Fig. S3, indicate that the stiffness coefficients ($K_{all}$) is identified as 12.5 N/mm ($R$-squared score: 0.9585).

$C[t]$ represents the baseline length of the artificial tendon, which is supposed to be a constant, but changes over time due to the downward migration ($\Delta L_1$) of the suit on the user body, caused by the reaction force from the cable sheath. This point can also be observed in Fig. S3, where the cable length retracted by the pulley increased over cycles, indicating that the overall length of the artificial tendon decreased due to the suit downward migration. Nonetheless, this migration tended to stabilize after the first 7-10 assisted GCs, and $\Delta L_1$ approximated a constant value. Here, $\Delta L_1$ is assumed to be constant within one GC and is estimated as follows:

$$\Delta L_1 = L_{meas}[t] - r \cdot \theta_{DF}[t] + \frac{F_{meas}[t]}{K_{all}} \tag{7}$$

where the subscript $_{meas}$ represents an actual measurement. This estimation is conducted once during the initial tightening of the actuation cable in each GC. Although the value of $L_1[t_0] + L_2[t_0]$ is a constant, it varies with different users' heights. It is confirmed at the initiation of the exoskeleton system, at which the cable will first be pre-tightened and then released to a safe position to allow the user to walk without any assistance in the following five GCs.

The overall length of the artificial tendon ($L$) can be calculated online based on the above analysis. This calculation provides feedforward information in following control system design to enhance the control performance.

**Model-Based Feedforward Profile Tracking Control**
Please note that ankle PF assistance is not required during the user's gait swing period; therefore, a transparent model, where the artificial tendon remains in a quasi-slack state and produces minimal external forces on the human body. Excessive cable push-out to achieve the quasi-slack state necessitates additional time for pre-tightening the cable at subsequent foot contact, which in turn reduces control responsiveness, particularly during early stance. Therefore, the length command of the artificial tendon in the slack state is iteratively adjusted to simultaneously minimize user interference and cable push-out, as follows:

$$L_{swing}[n+1] = L_{swing}[n] + \frac{F_{swing.max}[n] - 3}{K_{all}} \tag{8}$$

where $L_{swing}$ denotes the artificial tendon length desired to achieve a quasi-slack state, which remains constant within each GC, $n$ denotes the $n$th GC, $F_{swing,max}$ denotes the maximum cable force measured during swing, and $K_{all}$ is the coupled stiffness defined in Eq. (6). The desired maximum cable force during swing is set to 3 N in Eq. (8), and can be modified as needed. As the motor servo driver operates in a velocity mode, the artificial tendon length command obtained from Eq. (8) is transform into a velocity command using proportional-integral (PI) control with damping injection [44], as follows:

$$V_{swing,cmd} = K_p \cdot e_L + K_i \cdot \int_{t_0}^{t} e_L dt - K_d \cdot \dot{L}_{meas} \tag{9}$$

where $K_p$, $K_i$, and $K_d$ are gains, $e_L$ equals ($L_{swing} - L_{meas}$), where $L_{meas}$ is actual measurement, $t_0$ is initial time of the current GC, $\dot{L}_{meas}$ is determined by the pulley angular velocity and the radius of the pulley wheel.

A generic force control method for cable-driven exoskeletons typically maps force tracking errors to a cable length or velocity command, depending on the motor servo driver operation modes. As our driver operates in a velocity model, the force is mapped to a cable retraction speed as follows:

$$V_{FB} = \frac{F_{des} - F_{meas}}{Ms + B} \tag{10}$$

where $V_{FB}$ denotes the cable retraction speed corresponding to force error feedback ($F_{des}$ - $F_{meas}$), $F_{des}$ is the desired force calculated using Eq. (1), $F_{meas}$ is the actual force measured via load cells, and $M$ as well as $B$ are gains that determine the mapping dynamics.

However, the low mechanical bandwidth of soft exoskeletons limits the responsiveness of feedback control alone. Commonly used iterative learning control is unsuitable for non-steady locomotion due to the absence of clear and consistent periodicity. Inspired by the work [45], we further incorporate a feedforward component derived from the artificial tendon model in Eq. (6), as follows:

$$V_{stance,cmd} = V_{FB} - V_{FF} \tag{11}$$

where $V_{stance,cmd}$ is the velocity command during stance. $V_{FF}$ represents a feedforward component, as follows:

$$V_{FF} = \frac{dL[t|F_{des}(\theta)]}{dt} = r \cdot \frac{d\theta_{DF}}{dt} - \frac{1}{K_{all}} \cdot \frac{dF_{des}(\theta)}{dt} \tag{12}$$

where $L[t|F_{des}(\theta)]$ represents the artificial tendon length at a given desired force $F_{des}(\theta)$, $(d\theta_{DF}/dt)$ is the ankle DF angular velocity, which is measured in real time via IMUs, and $(dF_{des}(\theta)/dt)$ is given as follows:

$$\frac{dF_{des}(\theta)}{dt} = A \cdot \exp\left(-\frac{(\theta - \mu)^2}{2\sigma^2}\right) \cdot \left(-\frac{(\theta - \mu)}{\sigma^2} \cdot \frac{d\theta}{dt}\right) \tag{13}$$

where $\sigma$ represents either $\sigma_1$ or $\sigma_2$, depending on the value of the shank angle $(\theta)$, as described in Eq. (1). The term $d\theta/dt$ is the shank angular velocity provided by the shank IMU. The velocity command $(V_{stance,cmd})$ is fed into the motor servo drivers to implement the assistance profile tracking.

**Shank Angle-Based Exoskeleton Control System Design**

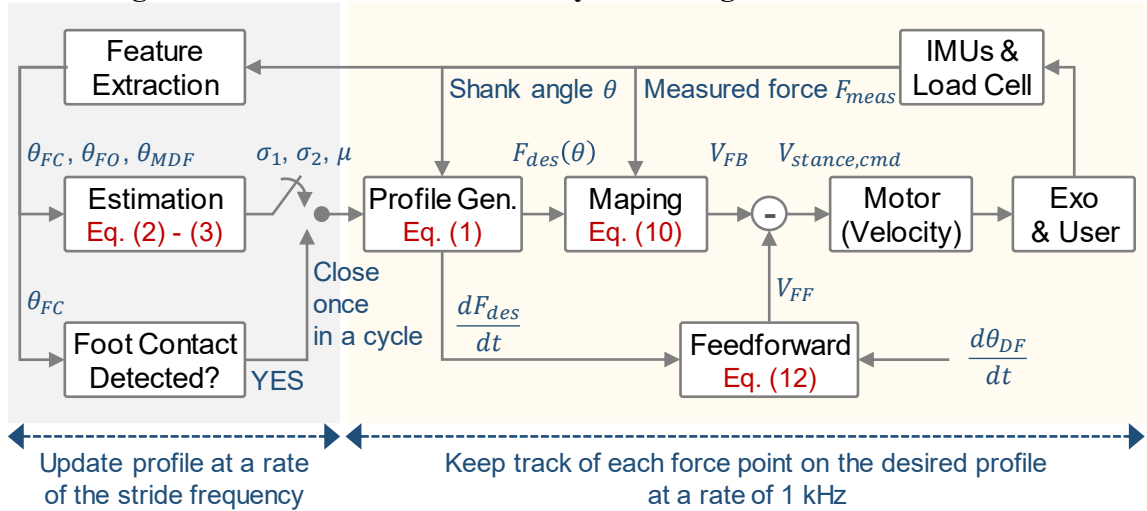

**Fig. 4.** Schematic block diagram of the proposed control system for the soft exoskeleton during stance phase. It consists of two parts that are operated at different frequencies. The first part focuses on generating a new desired assistance profile for the current GC, while the second part involves sampling assistive forces from the profile at each shank angle and controlling the exoskeleton to precisely apply the assistive forces on the user body.

As shown in Fig. 4, the designed control system consists of two parts that operate at different frequencies. The first part estimates a new set of Gaussian parameters $(\sigma_1, \sigma_2, \mu)$ based on human gait data $(\theta_{FC}, \theta_{FO}, \theta_{MDF})$. These parameters are transmitted to the second part at the start of the foot contact event. The second part generates the desire assistive force profile $(F_{des})$ based on the latest Gaussian parameters and samples the profile at each shank angle $(F_{des}(\theta))$. The model-based feedforward control method generates the velocity control command ($V_{stance,cmd}$) and applies the sampled force ($F_{des}(\theta)$) to the user's body through the exoskeleton's soft cable and suit. Concurrently, IMU sensors measure human gait data, which are collected and stored on the PCM3365 embedded controller for the Gaussian parameter estimation. Pseudocode for the control system implementation is provided in Supplementary Data File S2.

**Experimental Protocols**

*Experimental setup*

Various laboratory equipment and software were utilized in the experiments to collect data for analysis of subjects' metabolic rates, muscle activations, kinematics, and kinetics. An overall experimental scenario is presented in Fig. S4. The specific placement of markers and sEMG sensors on the human body is shown in Fig. S5. Detailed information about the equipment, data collection and processing pipelines, and participants' characteristics are provided in Supplementary Materials and Methods.

*Experiment design and parameter identification*

We conducted two single-subject experiments (session 1, 2) on the soft ankle exoskeleton platform, to validate the profile generation and tracking abilities of the proposed control system and the synchronization and biomimetic properties of the designed profile under non-steady conditions, respectively. A multi-subject experiment (session 3) was conducted to evaluate the effect of the exoskeleton assistance on human physiology and biomechanics during LW, LR, RA, and RD activities. The protocol and results of each experiment are detailed in following subsections. The values of parameters involved in the proposed control system are presented in Table S1. $A$ was experimentally set to 15% or 20% BW to avoid both discomfort and insufficient effects on human body. The initial values for Gaussian parameters ($\sigma_1$, $\sigma_2$, and $\mu$) were determined based on their averaged values across three subjects achieved in pre-tests. $r$ was individually measured. The coupled stiffness ($K_{all}$) was fitted experimentally as described in Eq. (6). The gains for the PI control with damping were results of manual tuning. First, $K_p$ was adjusted independently to achieve fast responsiveness. Next, $K_i$ was introduced to eliminate steady-state errors. Finally, $K_d$ was added to reduce overshoot. This tuning pipeline is effective and efficient for soft material made exoskeletons. The mapping gains ($M$ and $B$) depend on the pulley wheel radius, reduction ratio, and pattern characteristics of desired force profiles. Given the smoothness of the designed profiles, $M$ was set to zero, while $B$ was set to 15.7 N/mm based on experimental tuning.

*Protocol for session 1: functionality of each proposed module*

In session 1, we evaluated the proposed control system' functions of measuring user motion, regulating assistance profiles, and tracking the profiles. This experiment involved a single subject, who was instructed to wear the soft ankle exoskeleton on bilateral limbs and perform two minutes of each activity of LW (1.33 m/s), LR (2.20 m/s), RA (1.33 m/s; 10 deg), and RD (1.33 m/s; -10 deg) on a treadmill. The treadmill speed was determined by the subject and remained constant throughout each trial. Rest intervals of no less than five minutes were provided between activities. Moreover, an additional test was conducted on one subject with the treadmill belt speed changing, aiming to observe the adaptation process of the proposed profile and its model parameters ($\mu$, $\sigma_1$, and $\sigma_2$) to speed changes during the four types of activities, which can also serve as an experimental verification of the proposed control system during walking and running activity transitions. Detailed protocol for this test and results can be found in Supplementary Materials and Methods.

### Protocol for session 2: assistance adaptation to non-steady conditions

In session 2 (see Fig. 5A for protocol illustration), we evaluated the synchronization and biomimetic properties of the exoskeleton assistance profiles generated under non-steady conditions across the four types of activities. We adjusted the treadmill's bilateral belt speed online at a rate of 200 Hz using a Matlab program to enable controllable non-steady locomotion conditions. The treadmill acceleration was set as 25 m/s$^2$. As shown in Fig. S6, a forward or backward speed perturbation is applied as non-steady events, which first increases or decreases the belt speed by 80% within 0.1 s and then returned to its original speed within another 0.1 s. The onset of the perturbations was fixed at 15% GC, but the cycles in which the perturbation will be introduced and the order of applying the forward or backward perturbations were randomized. Belt speed command was sent 0.12 s ahead of the desired time from Matlab to the treadmill control software to compensate for time delays caused by signal transmission and the acceleration or deceleration of the treadmill belt. The desired, commanded, and actual (measured by the Vicon motion capture system) belt speed profiles are presented in Fig. S6. To facilitate the experiment, only the subject's left-side leg was configured with motion capture markers and analyzed for biomechanics. Moreover, we built a TCP/IP connection between the exoskeleton's on-board controller and the PC running the treadmill and motion capture software. The connection transmits the gait recognition results from the exoskeleton to the PC to ensure that the speed perturbations are always applied during the user's left-side gait stance phase. One subject selected 1.2 m/s for LW, RA, and RD testing trials, and 2.2 m/s for LR trial. Each trial lasted for sixty GCs and included four perturbations with two for forward and the another two for backward. Perturbations were restricted to one per GC and were not allowed to occur in consecutive cycles. A customized Matlab program for this protocol is provided in Supplementary Data File S3. A video clip presenting the treadmill belt speed perturbations is provided in Supplementary Movie S2.

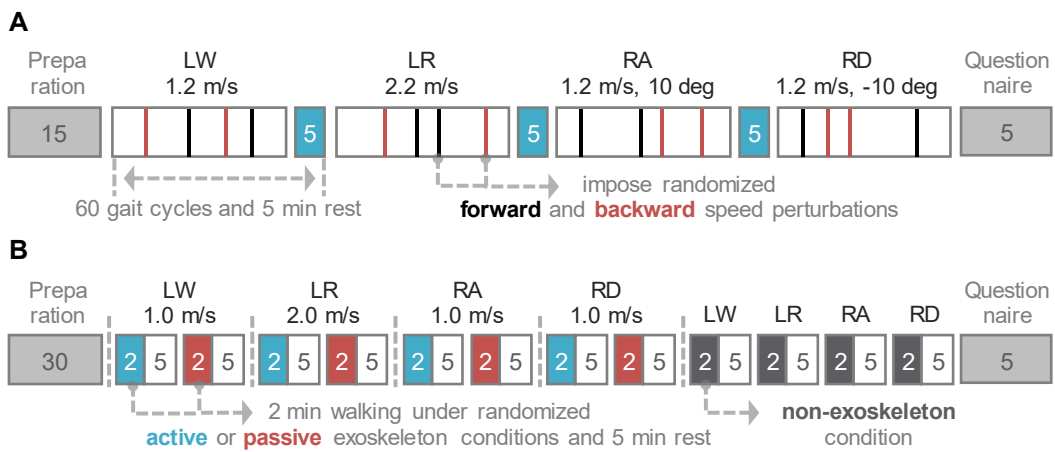

**Fig. 5.** Protocol for experiments. (A) Protocol for experiment session 2. Tests were conducted under non-steady locomotion conditions during LW, LR, RA, and RD activities. Forward or backward treadmill belt speed perturbations were imposed during the user gait stance phase to test the synchronization and biomimetic properties of the exoskeleton generated assistance profiles. (B) Protocol for experiment session 3. This experiment tested human biomechanical and physiological responses to the soft exoskeleton applying the proposed control system. Three exoskeleton conditions were applied.

### Protocol for session 3: human responses to exoskeleton assistance

In session 3 (see Fig. 5B for protocol illustration), we evaluated the effect of an active soft ankle exoskeleton applying the proposed control system on human physiology and biomechanics during the four types of activities. Eight subjects (Table S2) performed the activities under three conditions: active (ACT) and passive (PAS) exoskeleton conditions, in which the exoskeleton was worn on the subjects' bilateral limbs, and non-exoskeleton (NOE) condition. The mechanical assistance magnitude was set as 20% BW. The exoskeleton's ankle lever arm was measured manually for each subject before experiment. Each subject performed twelve 2-min locomotion bouts at a fixed treadmill speed of 1.0 m/s for level and ramp walking and 2.0 m/s for running. A 5-min sitting rest is included after each bout to let the leg muscles to recover from fatigue. The order of performing the four types of activities was fixed, while the order of the subjects walking under ACT and PAS conditions was randomized. All NOE bouts were conducted at the last to minimize experimental setup time, since exoskeleton suit and related modules, such as sensors and actuation cables, had to be removed from the body at this condition. A customized Matlab program for this protocol and a video clip demonstrating this experiment are provided in Supplementary Data File S3 and Movie S3, respectively.

**Statistical Analyses**

Statistical analyses were conducted in SPSS software with the significance level set at 0.05. In the experiments involving multiple participants and conditions (*experiment session 3*), after collecting data from all participants, we first confirmed the normality of the data using Shapiro-Wilk test. We then selected two-sided paired $t$-test for the analyses, or Wilcoxon Signed Rank test method when normality for the distributed data is not satisfied. We made our hypothesis that the soft exoskeleton applying the proposed shank angle-based control system would provide a positive effect on the user's physiology and biomechanics in diverse activities.

We calculated Pearson's correlation coefficient in SPSS software to indicate the strength of the correlation of the proposed assistance profile and the traditional time-based profile with the human biological ankle torques (*experiment session 2*). Coefficient results ($r$) were also reported. Linear fitting of the exoskeleton suit stiffness was conducted in Matlab using Curve Fitting Tool. Fitting result and its $R$-squared value are presented in Fig. S3. Moreover, unless otherwise specified, experimental results are presented in the form of mean $\pm$ standard deviation, where the standard deviation represents variations across GCs or participants.

# RESULTS

## Results of Session 1: Functionality of Each Proposed Module

The subject's foot pitch, shank, and DF angles measured by the exoskeleton's IMU sensors are presented in Fig. 6. These measurements are important for implementing the parameterization of profile progression and the estimation of the profile parameters. The model parameters converged to a set of relatively stable values after 9 to 11 GCs. The proposed updating method estimates the parameters directly based on the user's biomechanical characteristics, rather than relying on learning rules. Therefore, the convergence period is primarily attributed to the imposed limitation on the magnitude steps of parameter updates within each GC. Each component of the final motor velocity control command during the experiment, e.g., feedforward and feedback components, and tracking of the command are presented in Fig. S7. Corresponding desired and actual exoskeleton assistance force profiles for the four types of activities are presented in Fig. 6, using the percent GC and shank angle as the progression variables for the profiles. The root mean square error (RMSE) of the tracking in each cycle is calculated. Mean and standard deviation of ten GCs of the RMSE are calculated to indicate the control performance in the four types of activities as: $2.80 \pm 0.27\%$, $7.87 \pm 0.67\%$, $4.00 \pm 0.47\%$, and $5.27 \pm 0.27\%$, respectively, with respect to the maximum desired force (15% BW). Note that tracking errors appeared in the early tracking stage in the LR activity were due to the limited maximum speed of the motors, which resulted in the actuation cable not being fully engaged at the onset of profile tracking control. A video clip presenting the synchronized assistive force profile, GC, and video frames, and showing the subject walking at self-determined varying speeds, is provided in Supplementary Movie S4.

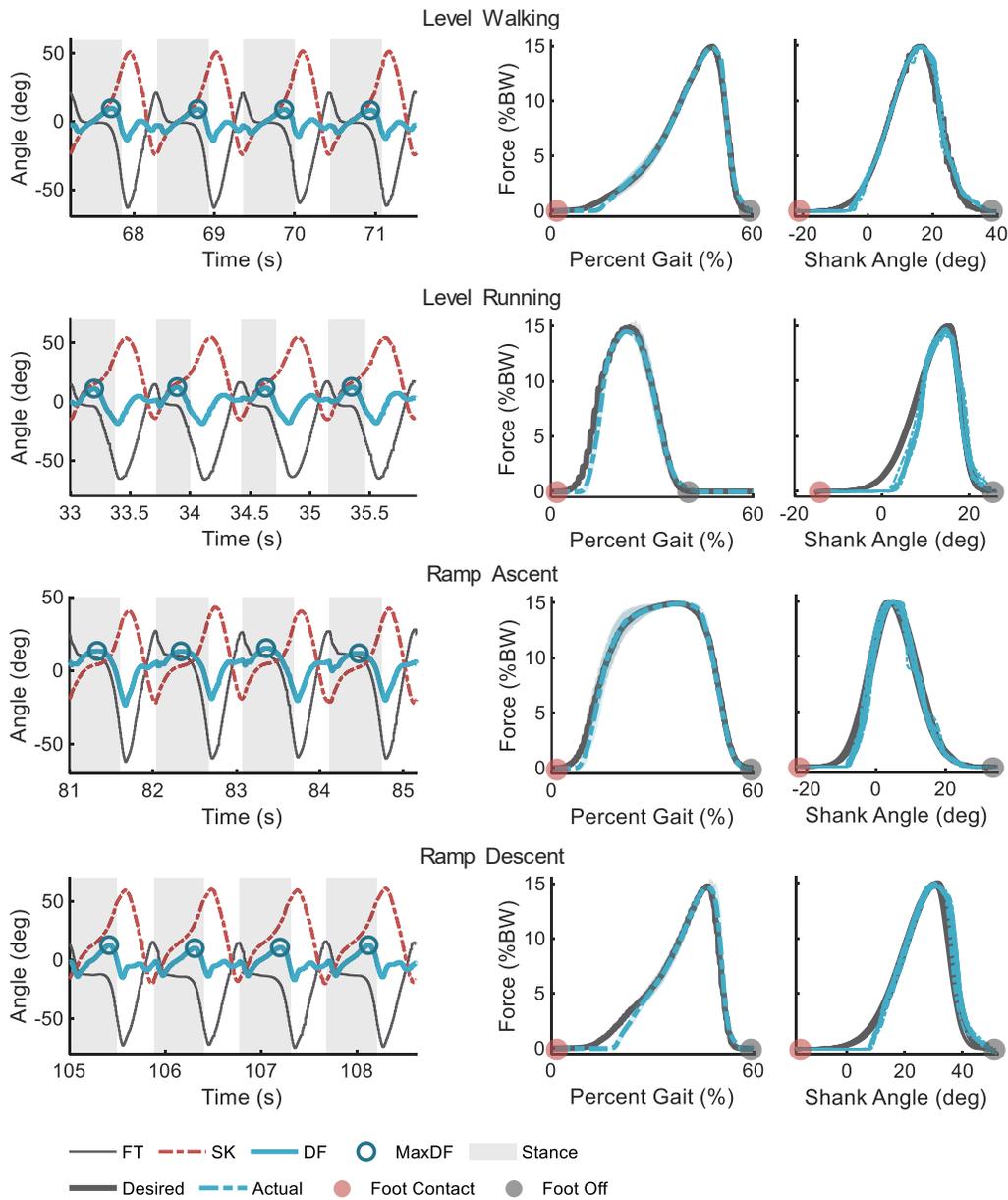

**Fig. 6.** Human motion data and exoskeleton's desired and actual assistance profiles collected in single-subject experiment. The left subfigures of each activity present four GCs of the subject's foot (FT), shank (SK), and ankle DF angles measured by the exoskeleton sensing modules. The right subfigures present the corresponding desired and actual exoskeleton assistance profiles with the percent GC and shank angle as the profile progression variable, respectively. The subfigure with the percent GC only aims to provide a more intuitive depiction of the assistance process. The shaded area represents the standard deviation across strides. For shank angle-based profile plots, no shaded area is shown, as these plots have dynamic desired profiles and progression variable ranges.

## Results of Session 2: Assistance Adaptation to Non-Steady Conditions

Experimental results are presented in Fig. 7. For comparison, we used existing methods to generate time-based exoskeleton assistance profiles for LW [46], LR [47], and RA [48] activities, except RD activity since we did not find a time-based profile generation method designed for assisting human ramp descending. The strength of the correlation between mechanical and biological profiles over stance phase is measured by Pearson's correlation coefficient ($r$). Correlation values for almost all shank angle-based profiles are greater than 0.90, with an average of 0.92. Time-based profiles exhibited acceptable biomimetic characteristics across the four types of activities under forward speed perturbations, with an average Pearson correlation of 0.86, but had remarkably decreased performance during backward perturbations, with an average Pearson correlation of 0.57. The results suggest that the proposed system enables exoskeletons to generate assistance profiles with a high level of correlation in patterns with human biological torques.

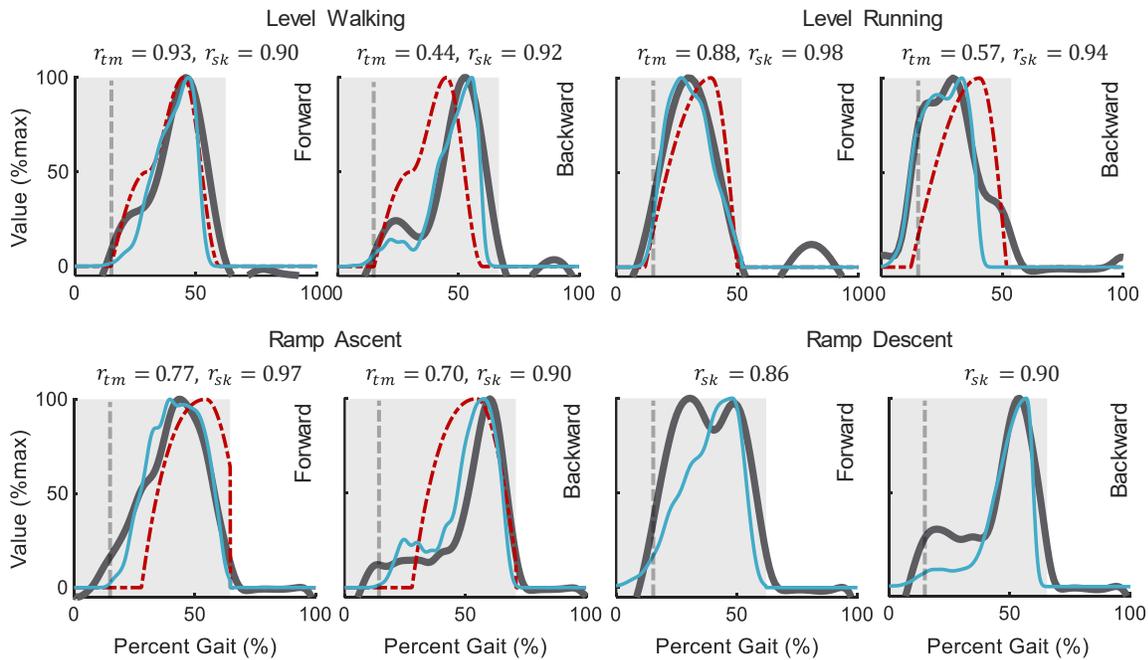

**Fig. 7.** Human ankle PF torques and assistance profiles. The left subfigure of each activity presents the biological torque and the mechanical force profiles of the subject performing activities under forward speed perturbations. The right presents the results under backward speed perturbations. $r_{tm}$ and $r_{sk}$ value represents the Pearson's correlation coefficient of the time- and shank angle-based profile versus the biological profile over the stance period. Magnitude of these profiles were normalized by their own maximum value for easy comparison.

## Results of Session 3: Human Responses to Exoskeleton Assistance
### Participants' kinematics and kinetics

Table S3 reports the exoskeleton dynamics. Table 1 reports the participants' kinematics and kinetics results across different conditions and activity types. Note that the trailing limb angle is defined as the angle in the sagittal plane at toe-off event, between the gravity vector and the vector from the fifth metatarsal joint to the great trochanter. Total moment or work rate refers to the combined contributions from both the human and exoskeleton. Assisted locomotion had reduced GC duration by around 4.12% (0.04 s) and 2.45% (0.03 s) in average compared with NOE and PAS conditions, respectively. It was observed that the external forces promoted users to lift their feet, and this happened more apparently in participants without prior exoskeleton experience. For example, veteran and novice users respectively reduced 2.85% ± 2.47% (0.03 s ± 0.03 s) and 7.41% ± 5.78% (0.08 s ± 0.06 s) in GC duration with respect to NOE condition during LW activity. Concurrently, the ankle angle range and trailing limb angle were also decreased when the participants walked with active assistance. Moreover, exoskeleton assistance did not produce a statistically significant effect on the participants' stance ratio, except the result of ACT versus PAS in RA activity ($p = 0.017$). Regarding kinetics, exoskeleton assistance reduced the peak total ankle PF moment in all types of activities compared with PAS condition (-3.56% [-0.06 Nm/kg] in average across activities) and in LR (-10.22% [-0.23 Nm/kg], $p = 0.012$) and RA (-0.77% [-0.01 Nm/kg], $p = 0.310$) activities compared with NOE condition. Peak total PF work rate in ACT had statistically significant differences with that in NOE (-14.25% [-0.46 W/kg] in average) and PAS (-15.44% [-0.49 W/kg] in average) across all the four types of activities. In summary, the results demonstrate that our active assistance is useful as it reduced participants' peak total ankle PF moment and work rate, but the burden from the passive exoskeleton needs to be further reduced to enhance the benefits. Additionally, participant kinematics were affected, which, although to some extent relies on the level of knowledge and familiarity of users with the exoskeleton and assistance strategy, is potentially associated with the improved kinetics.

**Table 1**
**Participant Biomechanics Results Across Conditions and Activities**

| Items | Unit | | Level Walking | | | Level Running | | | Ramp Ascent | | | Ramp Descent | |
|---|---|---|---|---|---|---|---|---|---|---|---|---|---|---|
| | | | NOE | PAS | ACT | NOE | PAS | ACT | NOE | PAS | ACT | NOE | PAS | ACT |
| Gait Cycle | s | Mean | 1.13 | 1.11 | 1.07 | 0.72 | 0.71 | 0.70 | 1.13 | 1.11 | 1.09 | 1.03 | 1.01 | 0.98 |
| | | SD | 0.03 | 0.07 | 0.08 | 0.04 | 0.04 | 0.04 | 0.05 | 0.08 | 0.08 | 0.07 | 0.07 | 0.06 |
| | | p-value | **0.015** | **0.023** | | 0.232 | 0.161 | | 0.052 | 0.472 | | **0.001** | **0.006** | |
| Stance Ratio | %GC | Mean | 67.39 | 67.33 | 67.31 | 51.35 | 50.42 | 50.80 | 68.32 | 68.98 | 68.10 | 69.08 | 68.91 | 67.93 |
| | | SD | 1.26 | 1.29 | 1.59 | 3.88 | 2.40 | 4.83 | 2.64 | 3.16 | 3.29 | 2.94 | 2.98 | 2.07 |
| | | p-value | <u>0.499</u> | 0.969 | | 0.632 | 0.795 | | 0.664 | **0.017** | | 0.313 | <u>0.575</u> | |
| Ankle Angle Range | deg | Mean | 24.65 | 23.98 | 18.77 | 36.19 | 35.21 | 32.73 | 30.81 | 30.53 | 27.52 | 22.99 | 23.65 | 17.23 |
| | | SD | 3.22 | 3.56 | 5.26 | 5.03 | 6.36 | 5.81 | 3.43 | 3.08 | 5.80 | 2.57 | 2.93 | 3.46 |
| | | p-value | **0.001** | **0.004** | | **0.008** | 0.063 | | **0.034** | 0.096 | | **0.025** | **0.007** | |
| Trailing Limb Angle | deg | Mean | 19.97 | 20.07 | 18.50 | 22.36 | 21.54 | 21.40 | 16.39 | 16.45 | 16.26 | 21.09 | 20.91 | 19.26 |
| | | SD | 1.76 | 2.42 | 2.98 | 3.22 | 3.19 | 3.56 | 2.30 | 2.31 | 2.14 | 2.91 | 3.40 | 3.62 |
| | | p-value | **0.046** | **0.004** | | 0.173 | 0.765 | | 0.797 | 0.735 | | **0.004** | **0.003** | |
| Peak Total PF Moment | Nm/kg | Mean | 1.26 | 1.29 | 1.26 | 2.25 | 2.19 | 2.02 | 1.30 | 1.32 | 1.29 | 1.05 | 1.07 | 1.05 |
| | | SD | 0.11 | 0.11 | 0.10 | 0.34 | 0.24 | 0.21 | 0.14 | 0.13 | 0.15 | 0.07 | 0.09 | 0.10 |
| | | p-value | 0.700 | 0.093 | | **<u>0.012</u>** | **0.010** | | <u>0.310</u> | 0.464 | | 0.812 | 0.407 | |
| Peak Total PF Work Rate | W/kg | Mean | 2.11 | 2.16 | 1.76 | 6.47 | 6.47 | 5.80 | 3.13 | 3.13 | 2.49 | 1.87 | 1.93 | 1.69 |
| | | SD | 0.38 | 0.39 | 0.36 | 1.21 | 1.17 | 0.93 | 0.54 | 0.53 | 0.36 | 0.48 | 0.42 | 0.61 |
| | | p-value | **0.001** | **0.009** | | **<u>0.012</u>** | **<u>0.012</u>** | | **0.017** | **0.020** | | **0.025** | **0.034** | |

p-value is obtained from the comparison versus the active (ACT) condition ($N = 8$, two-sided paired $t$-test if normality assumption is satisfied; otherwise, Wilcoxon Signed Rank Test is used and an underline is added). Bold signifies that the $p$-value is less than the pre-specified significance level (0.05).

### Participants' muscle activations

The percent changes of normalized sEMG root mean square (RMS) with respect to the results achieved in NOE are presented in Fig. 8A. Muscles showed different responses to the exoskeleton in the four types of activities. Overall, the passive exoskeleton did not impose a significant impact on human muscles when compared with the NOE condition, while assistance reduced human muscle activations, with the average changes of -7.50% (ACT) and -0.35% (PAS) in LW activity, -1.94% (ACT) and 1.97% (PAS) in LR activity, -2.06% (ACT) and 3.78% (PAS) in RA activity, and -6.33% (ACT) and 1.41% (PAS) in RD activity. The activation of lateral gastrocnemius (LG) muscle in ACT condition showed the most significant reductions across almost all activities, with reductions of -8.65% $\pm$ 3.05% in LW activity, -7.45% $\pm$ 1.44% in LR activity, -9.48% $\pm$ 1.90% in RA activity, and -13.64% $\pm$ 2.07% in RD activity, which is because the exoskeleton is intended to assist the user's ankle plantar flexors. In addition, rectus femoris (RF) muscle also showed reduced activation in all activities when assisted by the exoskeleton, since ankle PF is typically coordinated with hip flexion during locomotion. Thus, the assistance on ankle PF movement indirectly reduced the load on hip flexors, including the RF muscle. Moreover, the activation of biceps femoris (BF) and tibialis anterior (TA) muscles in PAS condition increased, especially in LR and RA activities, since their functioning stages were just the opposite of the exoskeleton assistance, and the mass of the exoskeleton added additional load on these muscles. Unexpectedly, LG muscle in all types of locomotion and RF in LW activity also experienced benefits in PAS condition. We had confirmed that no any assistive forces were applied during locomotion in PAS condition. The reduction may be due to compression from the kinesiology tape and exoskeleton suit. Regarding specific type of activities, all six muscles had reduced activations in LW activity. In the other three types of activities, while these muscles did not benefit as much as in LW activity, they still demonstrated promising performance, especially when compared with the results from PAS condition.

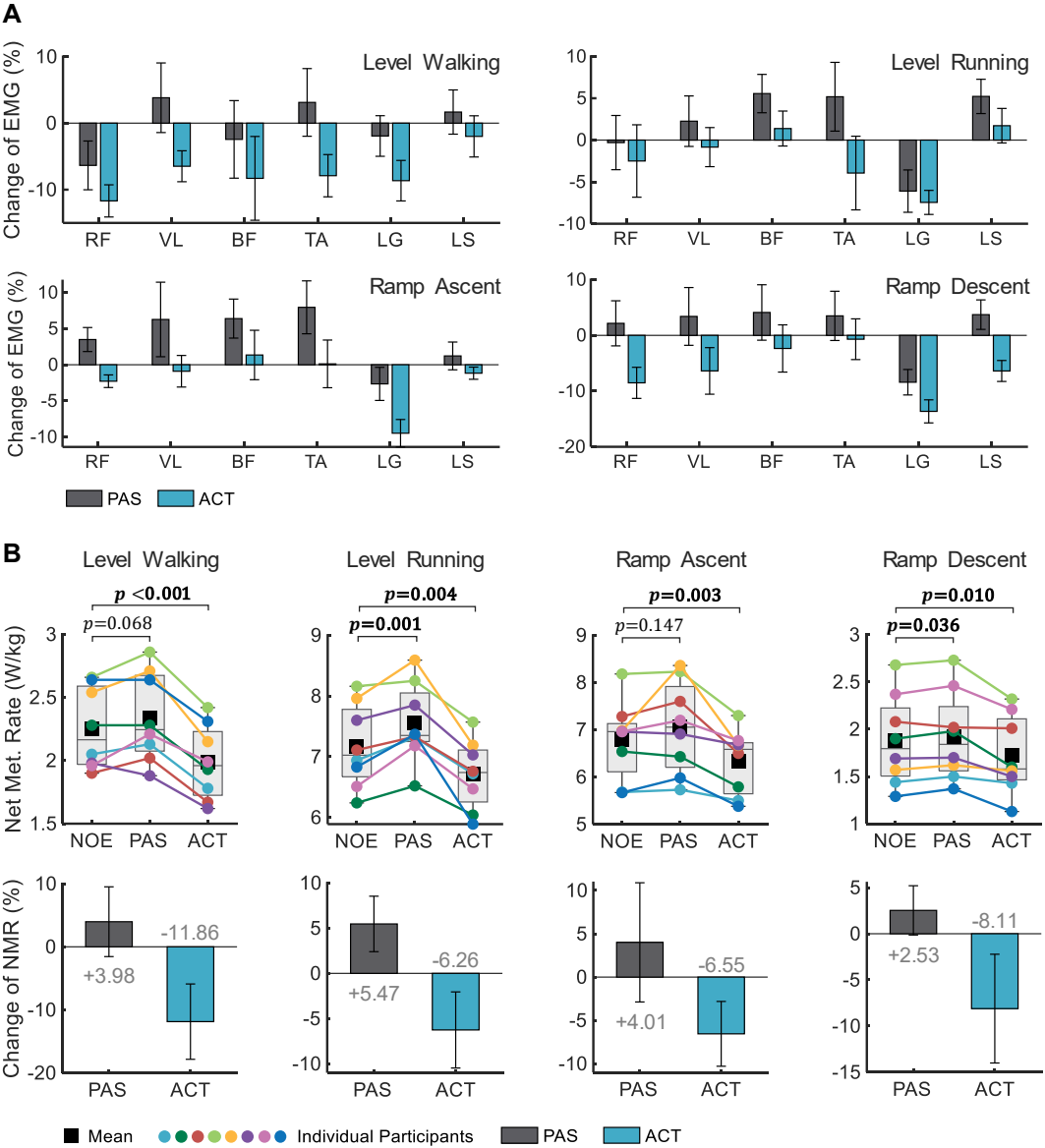

**Fig. 8.** Participants' physiological responses to exoskeleton assistance. (A) Changes of subjects' muscle activations. The subfigures present the improvement with respect to the results from NOE condition in the four types of activities. Negative value represents reduction. (B) Changes of subjects' net metabolic rates. The upper subfigures present individual participants' net metabolic rates (NMR) across NOE, PAS, and ACT conditions during each type of activities. The box plots and point plots show the distributions and trends. The $p$ values, calculated using the paired $t$-test method, indicate statistically significant differences compared with the NOE condition. The lower subfigures show the percent changes of the average net metabolic rates across participants under PAS or ACT conditions with respect to the NOE condition. The plus and minus signs before the values denote an increase and a decrease in percent change, respectively.

*Participants' net metabolic rates*

Individual participants' net metabolic rates under each condition and the percent changes of average net metabolic rates relative to the NOE condition are shown in Fig. 8B. The soft ankle exoskeleton affected users' metabolism during locomotion, both in active or passive conditions. Statistically significant improvements were observed in participants when receiving active exoskeleton assistance across all four types of activities, compared with the NOE condition, with especially notable reductions in net metabolic rates for LW and RD activities ($11.86\% \pm 5.97\%$ [$p < 0.001$] and $8.11\% \pm 5.89\%$ [$p = 0.010$] reductions, respectively). In contrast, participants received relatively small benefits from active exoskeleton in LR and RA activities, with net metabolic rate reductions of $6.26\% \pm 4.21\%$ ($p = 0.004$) and $6.55\% \pm 3.74\%$ ($p = 0.003$) respectively, compared with the NOE condition. The reduced benefits may be since in addition to ankle joints, humans rely on more hip joints during LR and RA. Passive exoskeleton increased the participants' net metabolic rates, with the highest increase observed in LR activity ($5.47\% \pm 3.07\%$ [$p = 0.001$]) compared with NOE. For LW, RA, and RD activities, net metabolic rates increased by $3.98\% \pm 5.53\%$ ($p = 0.068$), $4.01\% \pm 6.88\%$ ($p = 0.147$), $2.53\% \pm 2.67\%$ ($p = 0.036$), respectively. The wearable exoskeleton and sensors added less than 0.5 kg loads on the user's body. Thus, the increased metabolic rates in passive condition may be attributed to constraints imposed on user body movement by the soft exoskeleton suit, or varying familiarity levels among users with the exoskeleton system. Veteran users (P2, P4, P6) showed percent changes in net metabolic rates of 1.35% in PAS and -10.6% in ACT conditions, compared with NOE, while novice users experienced changes of 5.56% and -6.75%, respectively.

## DISCUSSION

### Time-Dependent Versus Time-Independent Exoskeleton Assistance Profiles

The ability to provide mechanical assistance that is insensitive to user gait perturbations is crucial for developing a time- and task-agnostic control system and advancing the practical application of assistive devices. Traditionally, the time variable is used to control the progression of exoskeleton assistance profiles, but it changes linearly and independently without any correlation with the human gait states. Consequently, any minor perturbations in human gait, such as early transitions between gait phases due to loss of balance, will cause mismatches in movement between the user and the exoskeleton. Even if the ideal mechanical assistance or biological profile is known, finding a simple, low-computation model to parameterize the profile over time remains challenging due to the shape distortions in local parts of the profile caused by the perturbations. Several definitions of phase variables have been proposed recently in literature, aiming to mitigate the mismatches. However, most of the phase variables are only capable of describing the progression of simple human and exoskeleton movements and ensuring their coordination, while the complex movement, such as walking backwards or others that involve backward progress of human movement, is not considered. With considerations of the mismatches and possible complex daily locomotion patterns, we propose using human shank angle to describe the progression of human gait and assistance profiles. This approach ensures consistent pairwise alignment between the user's gait states and the assistive forces, even under diverse conditions and activities. Most importantly, the designed shank angle-based assistance profile remains a significant correlation with human biological ankle torques, utilizing only a dual-Gaussian model and IMU measurements. These properties enable exoskeleton-generated forces to be applied to the user's body at appropriate time points while aligning with the user's biological characteristics. The idea of using human motion states as independent variables to regulate robot trajectories may inspire future studies aiming to build modal-adaptive, end-to-end, versatile control systems and improve coordination and interaction between robots and human users. In addition to provide bio-inspired mechanical forces, the dual-Gaussian piecewise function can also be used to control the delivered mechanical power or serve as a foundational model to embody optimization algorithms.

**Limitation and Future Works**

The present work has some limitations. First, only the profile's shape and progression are considered and adjusted online during assistance, while the magnitude is fixed to a percent value of the user's BW even in different activities. Foot pressure sensors can provide information about user's BW and the impact force from vertical acceleration, but do not account for user joint loads, as the loads also depend on joint motions. Moreover, incorporating pressure sensors imposes additional constraints on robot configurations, contradicting our goal of designing a simple, stable, and practical assistive robot. Another limitation is that we did not test our control system and exoskeleton device in real-life scenarios due to the off-board control box. To control extraneous variability in exploring and designing control systems, we opted to remove heavy modules from human body, as individuals exhibit different responses to additional weight.

Future studies should focus on enhancing the magnitude adaptation ability of the profile model and refining the exoskeleton device design to make it more comfortable and portable. Quantitative evaluation of the effect of exoskeleton assistance on human users remains a challenge in the field. The quality of sEMG and metabolic rate data is sensitive to environmental conditions and human physical states. Future studies should explore stable and consistent evaluation approaches for assistive devices to advance the field.

# CONCLUSION

In summary, this work presents a novel exoskeleton control system that uses human shank angle to describe the progression of human gait and assistance profiles, mitigating the movement mismatches between human users and the exoskeleton, especially under non-steady conditions. A dual-Gaussian profile model is designed using the shank angle variable, and a model parameter stride-by-stride online updating method is proposed based only on IMU measurements. This approach ensures the exoskeleton assistance profile consistently exhibit a bio-inspired similarity to human biological ankle torques across diverse activities. Additionally, a model-based feedforward control method is proposed, integrating the profile generation method to form a control system, which is implemented in a lightweight soft ankle exoskeleton device for experiments. Extensive experimental results demonstrated the effectiveness of the proposed control system in coping with non-steady conditions and the exoskeleton in assisting human locomotion. The present work may inspire future studies aimed at further improving movement coordination and cooperation between human users and robots.

# ACKNOWLEDGMENTS

**General**: We thank the members of the Robotics Laboratory at the Shenyang Institute of Automation, particularly Juhua Su, Peilin Yu, and Jiwei Li, for their valuable assistance in experiment preparation, data collections, and photography.

**Author contributions:** Conceptualization: XWT and WZJ. Methodology: XWT and WXC. Investigation: XWT, WZJ, and NL. Visualization: XWT. Funding acquisition: XWT, XGZ and BZ. Project administration: XGZ and YWZ. Supervision: XGZ and LQL. Writing – original draft: XWT. Writing – review & editing: XWT, BZ, NL, XGZ, and LQL.

**Funding:** This work was supported by the National Natural Science Foundation of China [grant numbers U22A2067, 62473361, 62403457]; the National Key Research and Development Program of China [grant number 2024YFB4708000]; the China Postdoctoral Science Foundation [grant number 2024M753412]; the Postdoctoral Fellowship Program of CPSF [grant number GZC20241852]; and the Fundamental Research Project of SIA [grant number 2024JC1K01].

**Competing interests:** The authors declare that they have no competing interests.

## DATA AVAILABILITY

Data to support the findings of this study are available from the first author upon request.

## SUPPLEMENTARY MATERIALS

Materials and Methods
Figures S1 to S12
Tables S1 to S3
Movies S1 to S4
Data Files S1 to S3

# Supplementary Materials for

**A Shank Angle-Based Control System Enables Soft Exoskeleton to Assist Human Non-Steady Locomotion**

Xiaowei Tan *et al.*

**The PDF file includes:**

Materials and Methods
Figures S1 to S12
Tables S1 to S3
Legends for movies S1 to S4
Legends for data files S1 to S3
Data file S1: SM_Data_File_S1.xlsx
Data file S2: SM_Data_File_S2.pdf
Data file S3: GitHub Link

For questions regarding the article or its supplementary materials, please contact Xiaowei Tan at tanxiaowei@sia.cn.

**Materials and Methods**

**Experiment Participants**

Approval for experiments involving human subjects was obtained from the People's Hospital of Liaoning Province Ethics Committee (2022HS007). Informed consent was secured from all subjects prior to their involvement. Only one subject participated in the first two experiment sessions, since the goal was to assess the functions of the designed control system and the correlation between the mechanical assistance and human biological moments, rather than examining the biomechanical responses of humans to the assistance. Besides, individual differences in behavioral and biomechanical responses to unexpected perturbations may introduce artifact and biases into the evaluation in the second experiment session. Eight subjects ($N = 8$, male; age = $25.8 \pm 3.1$ [23.0 - 33.0] years; mass = $71.9 \pm 5.2$ [68.0 - 80.0] kg; height = $1.75 \pm 0.04$ [1.72 - 1.82] m; mean $\pm$ standard deviation [range] unit) participated in the third experiment session, as reported in Table S2. Five of the subjects had no prior exoskeleton experiences and were not given additional time for acclimatization.

**Experimental Setup and Data Processing**

Energy expenditure of subjects was measured using an indirect calorimetry device (Oxycon Mobile, Vyaire Medical, Germany). Bread-by-breath gas data collected during the last one minutes of each trial were first cleaned by removing outliers based on the three-sigma rule and then processed [22] to calculate metabolic rates using the Brockway equation [49]. Net metabolic rates were derived by subtracting a resting metabolic rate from the overall metabolic rates, where the resting value was calculated from metabolic data collected while the subject was standing still. The net metabolic rates (W) was normalized by body mass (kg) to obtain comparable metabolic rates (W/kg). The average of the normalized metabolic rates for each subject was gathered to calculate the final indicator value corresponding to each testing condition, reported as mean $\pm$ standard deviation.

Surface electromyography (SEMG) signals were measured at 2 kHz using a wireless sEMG system (Ultium EMG, Noraxon, USA). As presented in Fig. S5, in addition to ankle plantar flexion (PF) muscles (*lateral gastrocnemius* [LG], *lateral soleus* [LS]), muscles involved in ankle (ankle dorsiflexion) DF (*tibialis anterior* [TA]), knee flexion (*biceps femoris* [BF]), and knee extension/hip flexion (*rectus femoris* [RF], *vastus lateralis* [VL]) were also examined to assess whether the robotic assistance desired for ankle PF assistance has an effect on the effort of muscles related to DF and other joints. Only the muscle activity in the left leg were measured. Medical PU film tape were applied over the sEMG sensors to prevent detachment due to skin sweating. MyoSync (Noraxon, USA) device was utilized to receive a sync signal from Lock Lab box (Vicon, UK) to trigger synchronized start or stop of the sEMG system with other systems. Raw signals were bandpass filtered (zero-lag 4th order Butterworth, cutoff 20-450 Hz), rectified, and lowpass filtered (zero-lag 4th order Butterworth, cutoff 6 Hz) using software (MR 3.16, Noraxon, USA) to create a linear envelop [50]. The envelope was segmented as individuals at each foot contact event identified using vertical ground reaction forces, and normalized to each gait cycle (GC). The average peak value of sEMG in baseline trials, where subjects walked without the exoskeleton, was calculated across ten GCs and used to normalized the amplitude of the corresponding muscle's sEMG envelope. The root mean square (RMS) value was calculated for each normalized envelope from ten consecutive GCs of data. Mean $\pm$ standard deviation was calculated from the RMS values of all individual envelopes from subjects to present the result for each condition.

A programmable split-belt force-sensing treadmill (Fully Instrumented Treadmill, Bertec, USA) was utilized, and the left- and right-side belts can be controlled independently including their speeds, directions, and acceleration via a TCP/IP connection. The treadmill measures the ground reaction forces on each belt, which are collected in Vicon software and used for later analysis of human biomechanics. The treadmill supported the simulation of non-steady locomotion conditions and the consistent implementation of experimental protocols involving multiple stages. A customized Matlab program for remote control of the treadmill is provided in Supplementary Data File S3.

Eight motion capture cameras (Vantage V5, Vicon, UK) were used to measure kinematic data of subjects. A customized marker set protocol was utilized, consisting of four markers for the pelvis, twelve markers for each leg, to capture human lower limb kinematic information, as presented in Fig. S5. Markers were intentionally placed not on the exoskeleton suit to avoid repositioning when donning or doffing the exoskeleton. Special attention was given to the Vicon software settings and treadmill baseline force calibration when changing the treadmill inclination. Medical PU film tape was used to attach markers on the skin, and kinesiology tape was applied to secure markers on fabric surfaces. These methods have been extensively verified in our laboratory for their effectiveness in securing markers stably even during vigorous movement and sweating. Vicon capture operations were controlled by a Matlab program (provided in Supplementary Data File S3) via a UDP connection, which was operated on the same PC station with the Vicon Nexus software. This program also managed the Bertec treadmill and exoskeleton system. Marker trajectories and the ground reaction forces from the treadmill were recorded at 100 Hz and 1000 Hz respectively and were time-aligned. The captured marker and force data were lowpass filtered in the Vicon Nexus software (zero-lag 4th order Butterworth, cutoff 10 Hz) and in Matlab (zero-lag 4th order Bessel, cutoff 12 Hz), respectively. Treadmill center of pressure data generated during the swing period were specially processed using linear interpolation in Matlab to mitigate signal jitters, which typically occur near the transitions between stance and swing phases.

Ten consecutive GCs of data with acceptable quality (without massive gaps and cross-steps on the belts) within the last one minutes were utilized to calculate the subjects' joint kinematics, moments, and work rates using inverse models in OpenSim software (a modified gait2354 model). Human analysis in OpenSim followed a standard pipeline: scaling the model dimensions using static data collected while each subject stood in a "T" pose; evaluating the position errors between actual and virtual markers; iteratively adjusting the positions of virtual markers; and performing inverse kinematics and inverse dynamics. Results from OpenSim were lowpass filtered (zero-lag 4th order Bessel, cutoff 8 Hz) before being reported in tables and figures. Moment (Nm) and work rate (W) data were normalized by each subject's body mass (kg) for comparison across subjects and conditions.

## Additional Experiment Under Slowly Changing Speed Conditions

An additional experiment was conducted on one subject with the belt speed slowly changing, aiming to observe how the proposed profile and its model parameters ($\mu$, $\sigma_1$, and $\sigma_2$) adapt to speed changes during the four types of activities. The treadmill belt speed was reduced by 50% with a 0.5 m/s$^2$ deceleration and then restored to its original speed with a 0.5 m/s$^2$ acceleration (see Fig. S8 for the protocol illustration). Each activity lasted for forty gait cycles (GCs). The GCs during which the speed changes occurred were randomized, with a minimum interval of five GCs between them. A 5-min rest period was allowed after each activity.

Human biomechanical data, shank angle-based profiles, and profile parameter estimations were recorded simultaneously (see Figs. S9 - S12 for the results). The biological ankle torques in the figures were first normalized by the maximum value across the entire trial (%max). We then adjusted the magnitude of the proposed assistance profile to match the magnitude of the current cycle's normalized biological ankle torques. This is because that we did not focus on the assistance magnitude parameter in the present work. Curves and points highlighted in red color indicate that they were generated during the period between the cycle starting to decease speed and the cycle starting to restore speed (including the two end cycles). Note that the Gaussian parameters were updated one-GC behind. The initial values for $\mu$, $\sigma_1$, and $\sigma_2$ were 15, 10, and 5, respectively.

As shown in Figs. S9 - S12, the values of the three parameters change noticeably with the human gait variations caused by speed changes. For example, a reduction of walking speed led to an extended gait stance phase during level walking, resulting in a delayed timing of peak ankle PF torque. The experimental results have demonstrated two points: 1) the profile model parameters can be updated step-by-step using the proposed updating method with only inertial measurement unit (IMU) measurements; and 2) the assistance profile generated by the proposed control system can keep aligned with the pattern of human biological torques, even under speed changing conditions.

Moreover, the experiment with speed reduction during level running can also serve as a verification of the proposed control system during activity transitions between running and walking. We also conducted experiments involving transitions between all the four types of activities. However, we found that the ground reaction forces measured by the Bertec instrumented treadmill are highly sensitive to changes in treadmill inclination and cannot be calibrated online, resulting in invalid ground reaction force data. Therefore, results of experiments involving online changes in treadmill inclination were not reported in the present study.

The results show that the pattern and magnitude of human biological ankle torques changed with locomotion speed. The three profile parameters updated stepwise in the exoskeleton on-board controller, modulating the generated profiles to continuously align with the patterns and characteristics of human ankle biomechanics and address the user's biological needs.

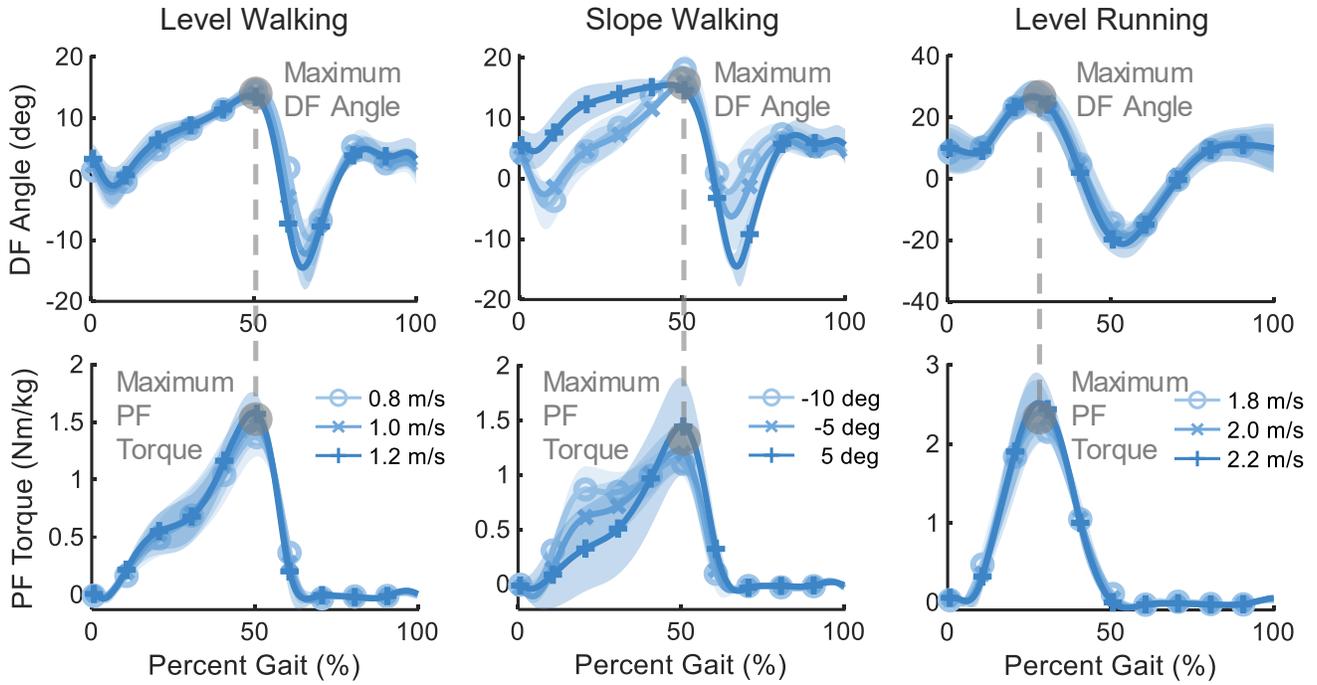

**Fig. S1. Human ankle DF angles and PF torques.** The dataset was collected with the subjects walking and running on a treadmill without inclination. Slope walking is performed at a speed of 1.0 m/s. The line and the shaded area indicate the average and the standard deviation across subjects.

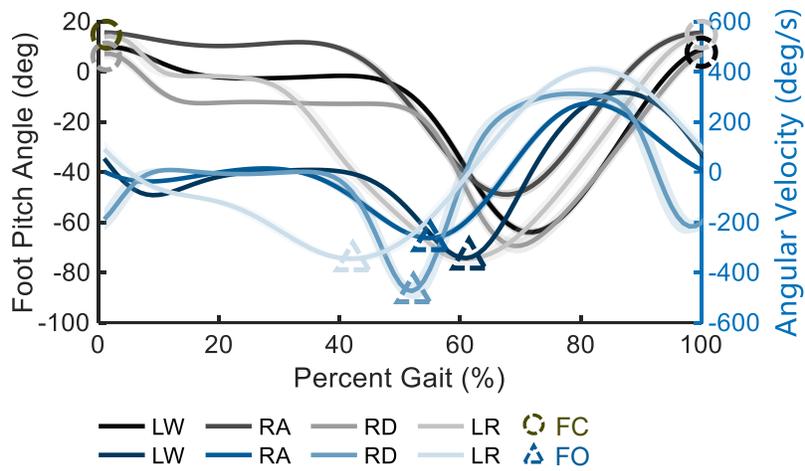

**Fig. S2. Human foot pitch angle (black) and angular velocity (blue) curves.** The motion data were collected in level walking (LW; 1.0 m/s), ramp ascent (RA; 1.0 m/s; 10 deg), ramp descent (RD; 1.0 m/s; -10 deg), and level running (LR; 2.0 m/s) activities. The shaded area represents one standard deviation across GCs. "FC" and "FO" denote foot contact and foot-off, respectively.

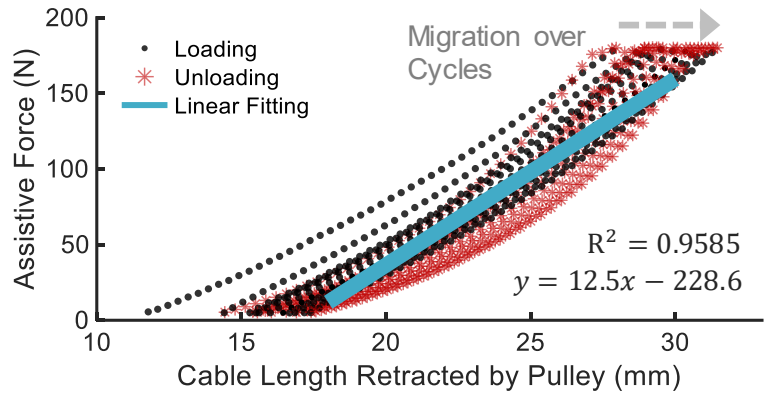

**Fig. S3. Human-exoskeleton coupled stiffness fitting.** Data points of assistive forces versus cable length recorded during ten cycles of the force loading (from 5 N to 180 N) and unloading (from 180 N to 5 N) processes. $R$-squared score for the linear fitting result is 0.9585.

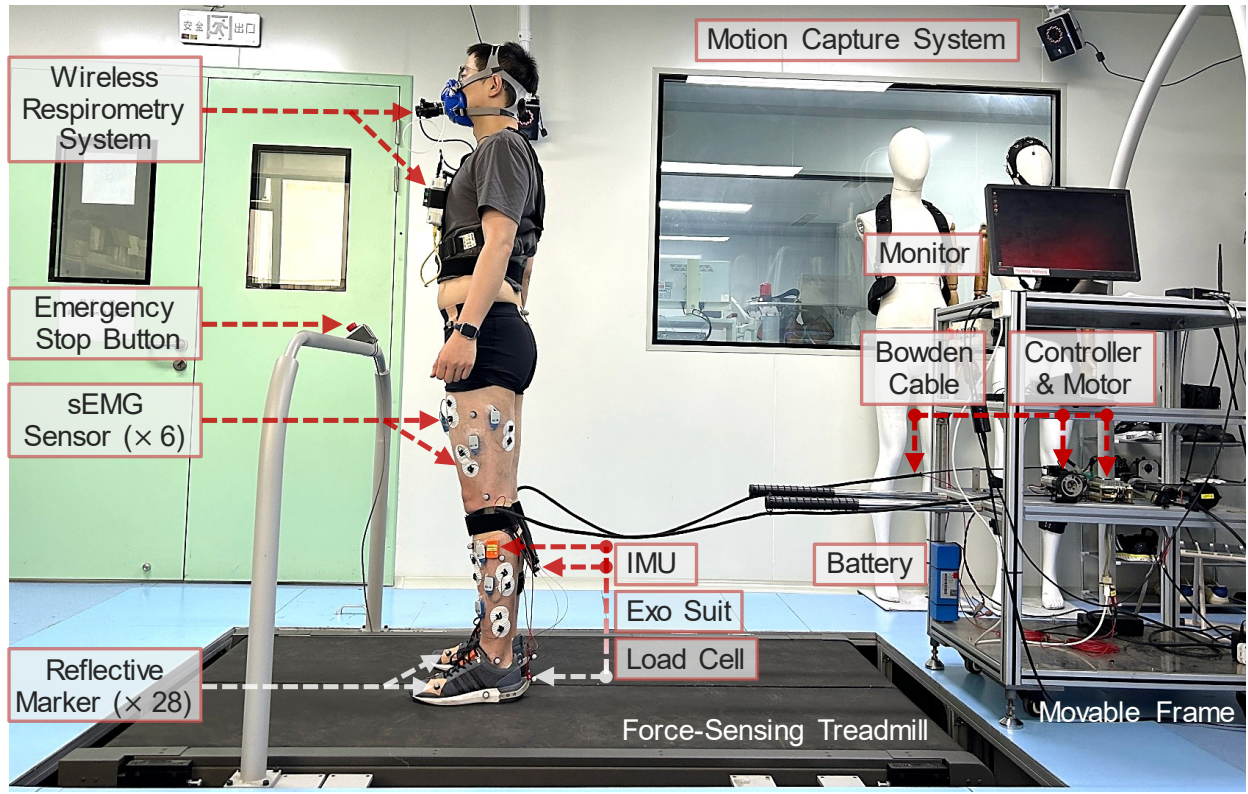

**Fig. S4. Photograph of the overall experimental scenario.** The systems used to evaluate the designed exoskeleton include a wireless respirometry system, a motion capture system, a sEMG measurement system, and a force-sensing treadmill. The exoskeleton's on-body components include wireless IMU sensors, wearable soft suit, and load cells. The off-body components include Bowden cables, a controller, two sets of motors and drivers, which are mounted on a metal frame positioned behind the treadmill. A desktop computer, positioned near the metal frame, is used to synchronously collect data from these systems.

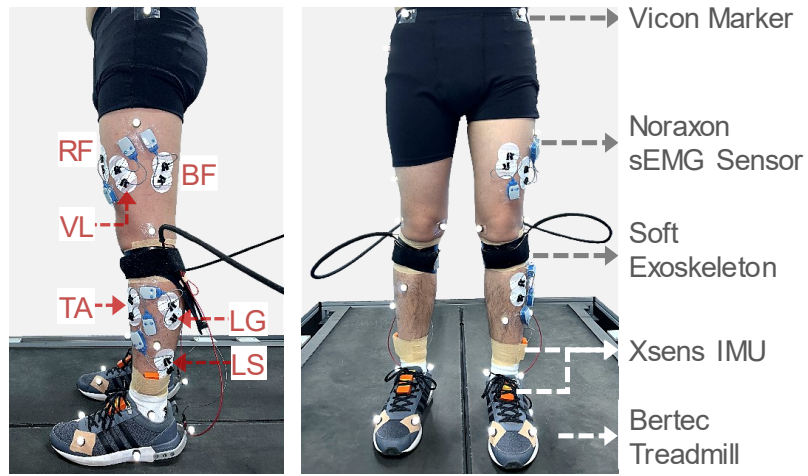

**Fig. S5. Experimental setup for muscle activity measurement and marker points.** Muscle activities were measured for *lateral gastrocnemius* (LG), *lateral soleus* (LS), *tibialis anterior* (TA), *biceps femoris* (BF), *rectus femoris* (RF), and *vastus lateralis* (VL).

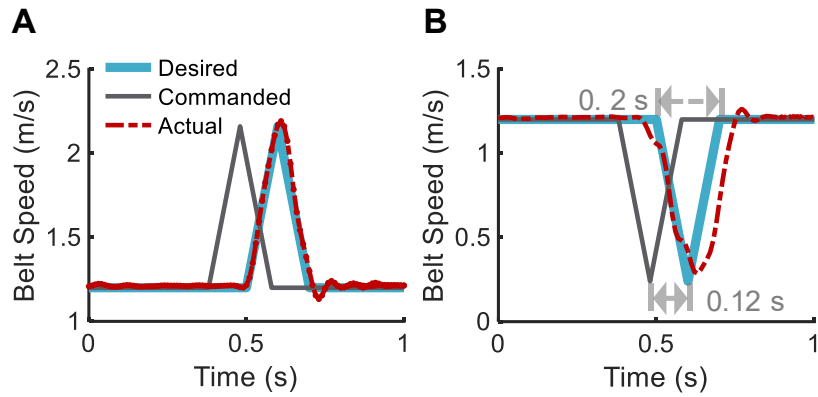

**Fig. S6. Treadmill belt speed profiles with perturbations.** (A) Desired, commanded (sent as a command to the treadmill speed controller), and actual (measured by the Vicon motion capture system) treadmill belt speed during forward speed perturbation. The commanded profile is set 0.12 s ahead of the desired profile to offset the time delays between the desired and the actual profiles. (B) Treadmill speed profile during backward speed perturbation.

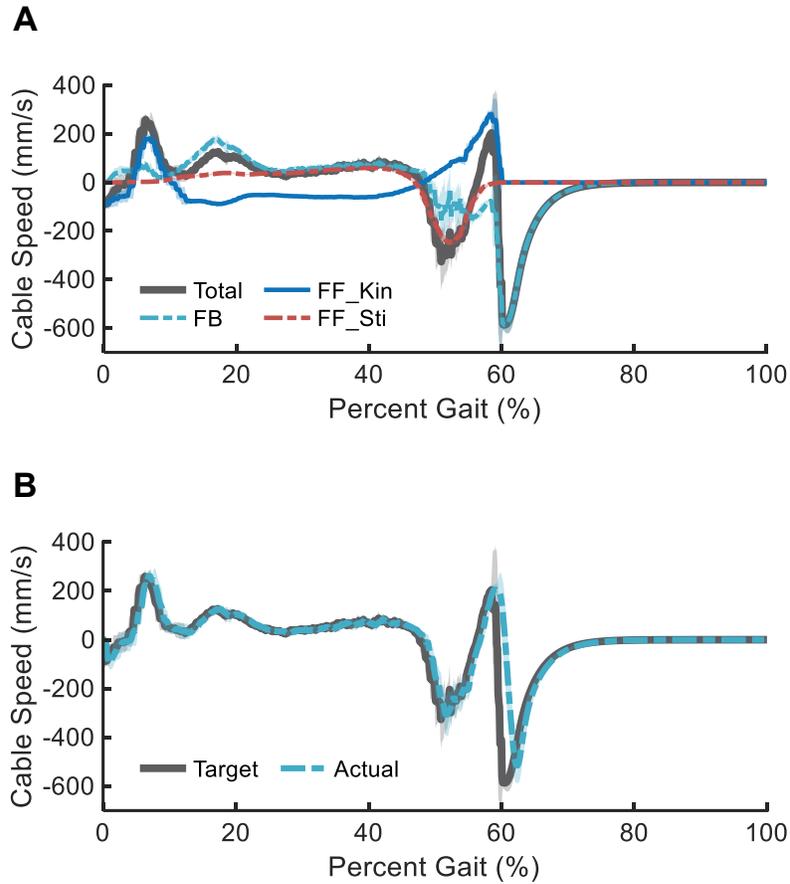

**Fig. S7. Tracking control of actuation cable retraction speed.** (A) Components of the total (final) speed command. The total command is generated to retract the actuation cable and produce desired assistive forces on the user's body. As presented in Eqs. (11) - (12), it consists of one feedback (FB) component and two feedforward (FF) components that are derived from the human-exoskeleton kinematics (FF_kin) and stiffness (FF_Sti) models, respectively. (B) Tracking of the total (target) speed command. The shaded area represents the standard deviation across ten strides.

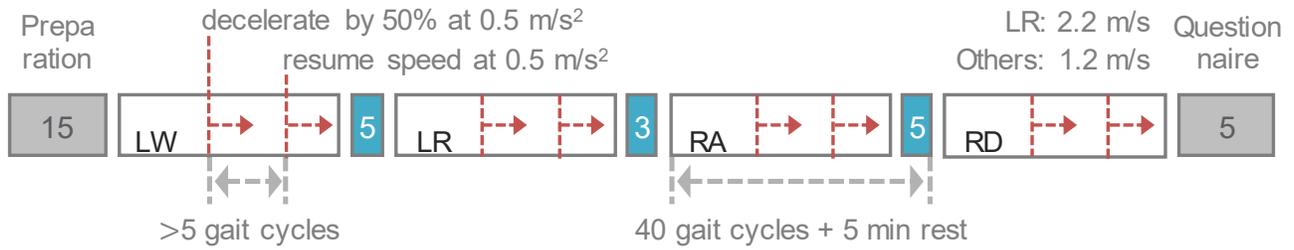

**Fig. S8. Experimental protocol.** It describes the detailed protocol for the experiment conducted under treadmill belt speed slowly changing conditions. Treadmill belt speed was changed by 50% at a 0.5 m/s² acceleration, and then returned to the original speed after an interval of no less than five GCs. A 5-min rest was allowed after each locomotion bout. Ramp activities were performed with the treadmill inclination of 10 degrees.

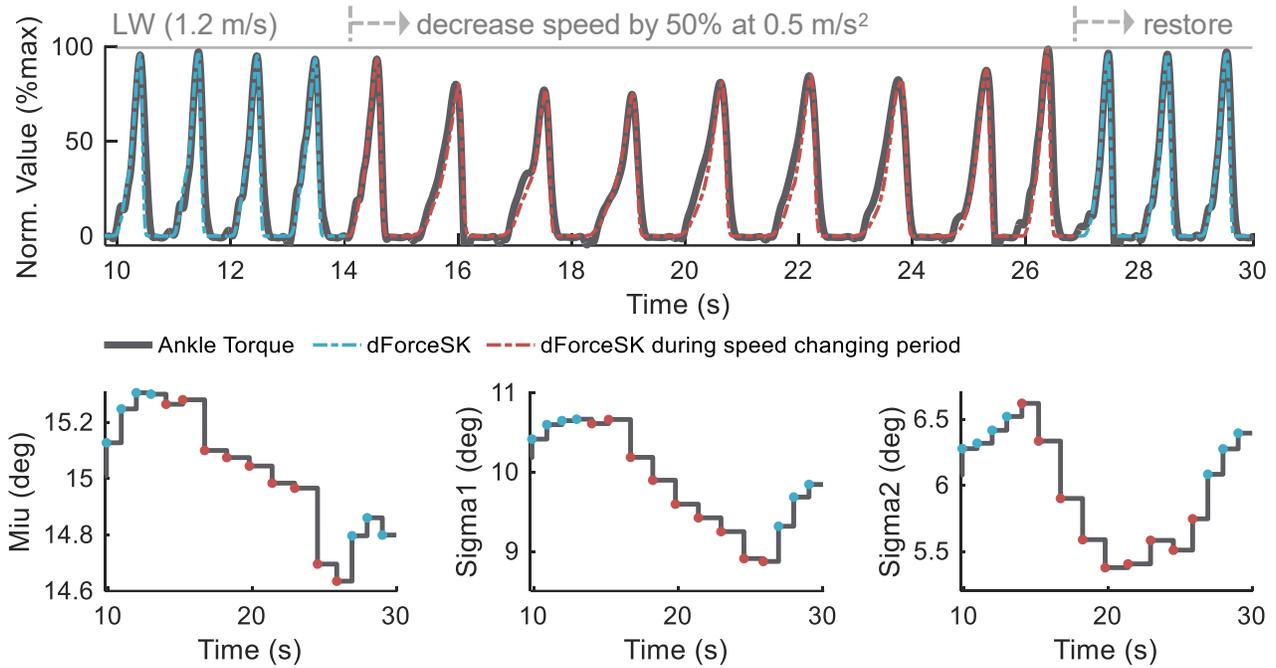

**Fig. S9. Biological ankle torques and exoskeleton assistance profiles during level walking (LW) activity.** The data were collected under treadmill belt speed slowly changing conditions. The bottom three subfigures are the profile model parameter estimations recorded simultaneously. Treadmill belt speed started to decrease at around 14.0 s and to restore at around 25.7 s. The red color highlights the period between the cycles where speed started to change.

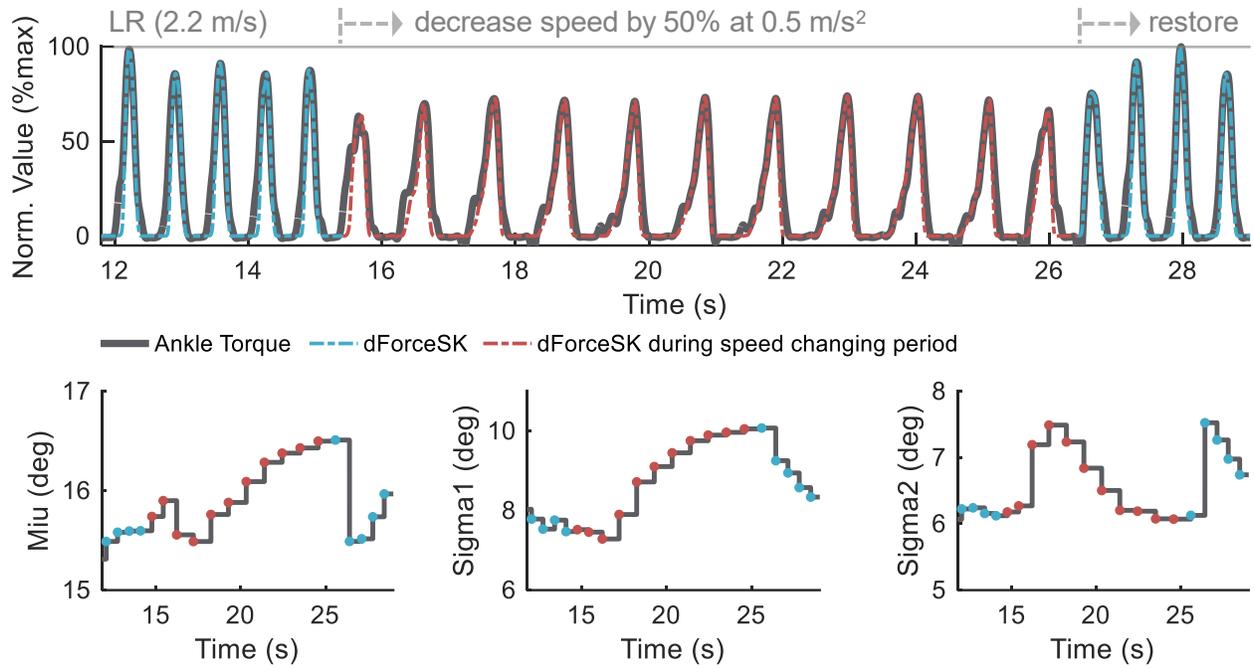

**Fig. S10. Biological ankle torques and exoskeleton assistance profiles during level running (LR) activity.** Treadmill belt speed started to decrease at around 15.0 s and to restore at around 25.5 s.

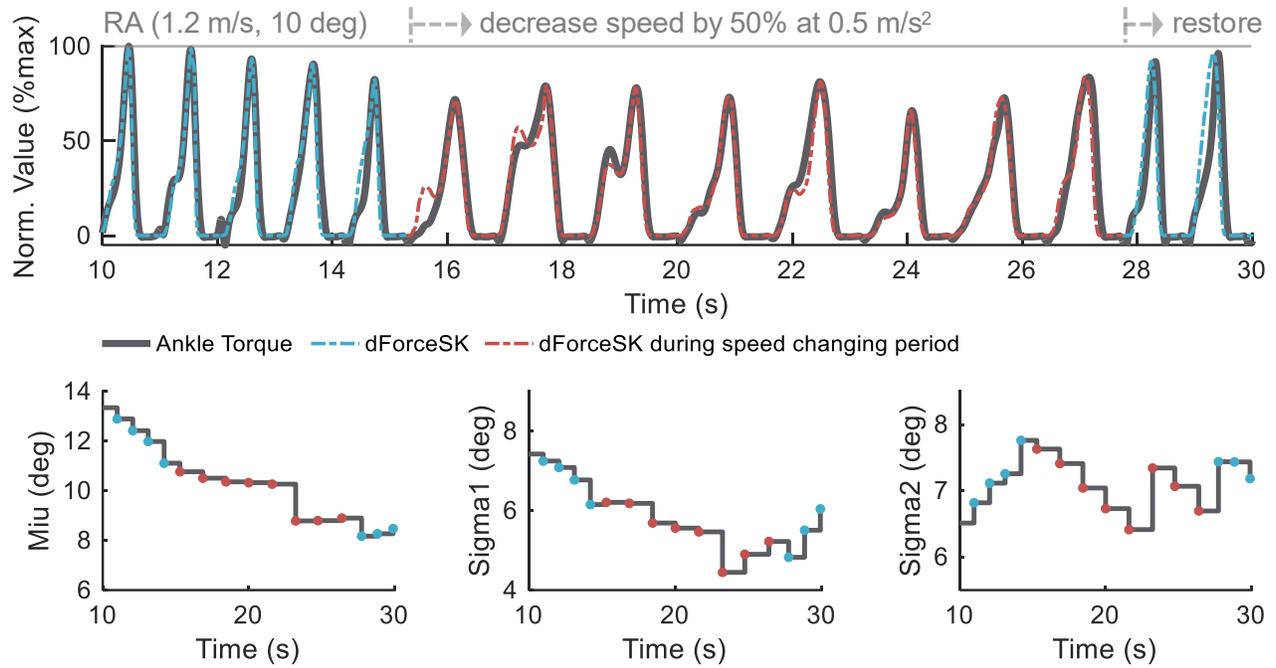

**Fig. S11. Biological ankle torques and exoskeleton assistance profiles during ramp ascent (RA) activity.** Treadmill belt speed started to decrease at around 15.3 s and to restore at around 26.3 s.

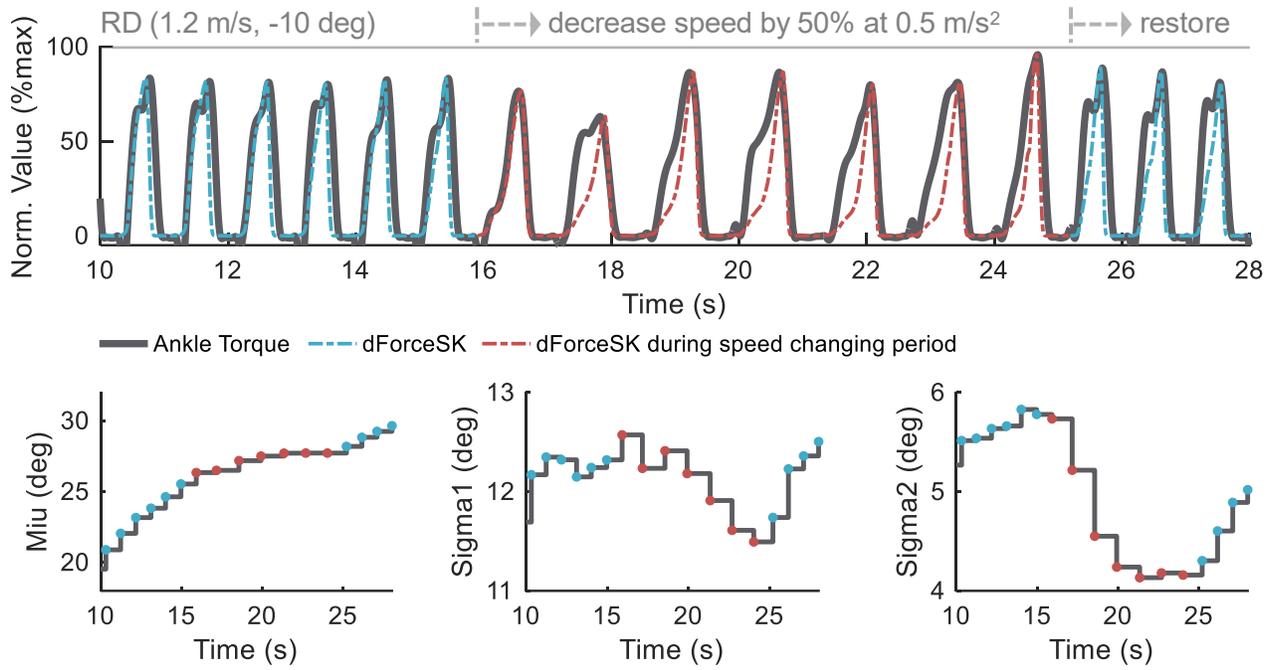

**Fig. S12. Biological ankle torques and exoskeleton assistance profiles during ramp descent (RD) activity.** Treadmill belt speed started to decrease at around 15.9 s and to restore at around 24.0 s.

**Table S1**
**Parameter Values of the Control System in Experiments**

| Para. | Value | Unit | Para. | Value | Unit |
|---|---|---|---|---|---|
| $A$ | 15% or 20% | BW | $K_p$ | 23 | 1/s |
| $\sigma_{1,initial}$ | 10 | deg | $K_i$ | 0.0001 | 1/s$^2$ |
| $\sigma_{2,initial}$ | 5 | deg | $K_d$ | 1.8 | - |
| $\mu_{initial}$ | 15 | deg | $M$ | 0 | N/(s·mm) |
| $r$ | subject-specific | | $B$ | 15.7 | N/mm |
| $K_{all}$ | 12.5 | N/mm | | | |

**Table S2**
**Experimental Participant Characteristics**

| | Sex (-) | Age (years) | Mass (kg) | Height (m) | Foot Size (mm) | Lever Arm (mm) | Exoskeleton Experience |
|---|---|---|---|---|---|---|---|
| P1 | M | 26 | 74 | 1.74 | 260 | 100 | NO |
| P2 | M | 23 | 75 | 1.73 | 260 | 90 | YES |
| P3 | M | 26 | 80 | 1.82 | 265 | 110 | NO |
| P4 | M | 25 | 70 | 1.73 | 260 | 90 | YES |
| P5 | M | 24 | 68 | 1.77 | 265 | 95 | NO |
| P6 | M | 33 | 70 | 1.80 | 265 | 110 | YES |
| P7 | M | 25 | 70 | 1.72 | 260 | 85 | NO |
| P8 | M | 24 | 68 | 1.72 | 260 | 85 | NO |

**Table S3**
**Exoskeleton-Delivered Peak Mechanical Moment and Work Rate**

| Items | Unit | | LW | RN | RA | RD |
|---|---|---|---|---|---|---|
| Peak Mechanical PF Moment | Nm/kg | Mean | 0.19 | 0.18 | 0.19 | 0.19 |
| | | SD | 0.02 | 0.02 | 0.03 | 0.02 |
| Peak Mechanical PF Work Rate | W/kg | Mean | 0.19 | 0.41 | 0.20 | 0.25 |
| | | SD | 0.03 | 0.13 | 0.03 | 0.08 |

**Movie S1. Soft ankle exoskeleton donning and doffing.** This movie shows one participant donning the exoskeleton platform's soft wearable parts in 24 seconds, and doffing them in 10 seconds.

**Movie S2. Forward and backward treadmill speed perturbations.** This movie shows a participant level and ramp walking on a treadmill at 1.2 m/s. Treadmill belt speed perturbations are imposed during both activities: a forward perturbation (increase the belt speed by 80% within 0.1 s) and a backward perturbation (decrease the belt speed by 80% within 0.1 s). These abrupt speed changes simulate human non-steady locomotion conditions and are implemented via a remote treadmill speed control Matlab program (provided in supplementary data file S3).

**Movie S3. Exoskeleton experiments with the full setup.** This movie shows the overall on-body sensors and devices used in exoskeleton-assisted human experiments and a representative participant performing four types of activities: level walking, level running, ramp ascent, and ramp descent, under three conditions: exoskeleton-active, exoskeleton-passive, and without the exoskeleton.

**Movie S4. Exoskeleton-assisted walking at user-determined varying speeds.** This movie shows synchronized exoskeleton assistive force profiles, gait cycle progression, and video frames of a participant walking on a treadmill at varying speeds. The participant changes the walking speed, as indicated in the gait cycle progression line (grey), where changes in gait periods are reflected. The exoskeleton applying the proposed control system modulates its assistive profiles and delivers the assistance to the user body precisely.

**Data File S1. A bill of materials for the soft ankle exoskeleton platform.** This file provides the quantity, model, and manufacturer details for each component of the exoskeleton platform.

**Data File S2. Pseudocode for the proposed control system implementation.** This file provides the workflow of implementing the proposed control system, which consists of one main thread and three child threads (each for IMU signal processing, load cell measurement, as well as exoskeleton control).

**Data File S3. Matlab program for the experimental protocols.** Please note that this file was uploaded on GitHub, and the repository link is https://github.com/xw-TAN/SK_Exp_Protocol.This file provides the Matlab programs for the experiment session 2 and session 3 as described in the main content. Remote control for Bertec treadmill belt's speed, direction, acceleration and the implementation of belt speed perturbations can also be found in this file.

| | Component Description | Quantity | Model | Manufacturer |
|---|---|---|---|---|
| **On-body** | BOA Lacing System | 2 | L6 | BOA, (Produced in) China |
| | Magnetic Buckle | 2 | HOOK 25 | Fidlock, Germany |
| | Cable Clamp | 2 | C-Shape | Lebycle, China |
| | Push-to-Connect Fitting | 4 | PC5-M5C | Zhuoji, China |
| | IMU | 2 | Awinda | Xsens, Netherlands |
| | Load Cell | 2 | LSB205 | Futek, USA |
| | Shoe | 1 pair | ART AW5271 | Adidas, (Produced in) Vietnam |
| | Wearable Suit | 2 | N/A | Customized |
| **Off-body** | Embedded Controller | 1 | PCM-3365 | Advantech, China |
| | Power Module | 1 | PCM-3910 | Advantech, China |
| | Load Cell Amplifier | 2 | USB220 | Futek, USA |
| | EC-4pole Motor | 2 | 305013 | Maxon, USA |
| | 66:1 Planetary Gearhead | 2 | 326665 | Maxon, USA |
| | ENC 16 EASY Encoder | 2 | 499361 | Maxon, USA |
| | Motor Driver | 2 | Gold Solo Twitter | Elmo Motion Control, Israel |
| | Battery | 1 | 11.6Ah | Jinfeng, China |
| | Cable Sheath Holder | 2 | 10H | Shuiniu, China |
| | Bowden Cable Sheath | 3 m | BHL 100 | Jagwire, China |
| | Bowden Cable | 4 m | 1.5 mm | Lebycle, China |
| | Pulley Cartridge | 2 | N/A | Customized |
| **Other Devices** | Split-Belt Treadmill | 1 | N/A | Bertec, USA |
| | Motion Capture Device | 1 | Vantage V5 | Vicon, UK |
| | Portable Respirometer | 1 | Oxycon Mobile | Vyaire Medical, USA |
| | EMG Measurement Device | 1 | Ultium EMG | Noraxon, USA |
| | Synchronization Device | 2 | Lock Lab/ MyoSync | Vicon, UK/ Noraxon, USA |

Article Title:     A Shank Angle-Based Control System Enables Soft Exoskeleton to Assist Human Non-Steady Locomotion

---

**Main Thread:** Exoskeleton Program Entry - Create and monitor child threads

---

**Output:** flagToRunIMUThread, flagToRunForceThread, flagToRunRobotThread

1    Initialization, create a log file, and set the child thread flags with TRUE
2    Create IMUSensor, ForceSensor, RobotControl child threads
3    Code block here until all child threads finished
4    Deinitialization

---

**Child Thread 1:** IMUSensor - Process IMU measurements and estimate Gaussian profile parameters

---

**Input:** flagToRunIMUThread

**Output:** $\theta_{SK}$: shank angle [Unit: deg]
          $\dot{\theta}_{SK}$: shank angular velocity [Unit: deg/s]
          $\theta_{DF}$: ankle dorsiflexion (DF) angle [Unit: deg]
          $\dot{\theta}_{DF}$: angle DF angular velocity [Unit: deg/s]
          $S_{pha}$: indicate foot contact or foot-off event
          $\vec{G}(n) = (\sigma_1, \sigma_2, \mu)$: Gaussian parameters estimated at $n$th gait cycle

1    Initialization
     */\* create a child thread for each leg (two IMUs), each thread performs the following codes    \*/*
2    Get initial shank and foot angles as the user stands straight for later calibration
     */\* while loop frequency: 100 Hz    \*/*
3    **while** *flagToRunIMUThread* **do**
4       Check data availability and synchronization of the two IMUs
5       Update $\theta_{FT}, \theta_{SK}, \dot{\theta}_{FT}, \dot{\theta}_{SK}$   */\* $\theta_{FT}$ is positive as the forefoot is vertically higher than the hindfoot;*
         *$\theta_{SK}$ is positive as the ankle joint is horizontally behind the knee joint    \*/*
6       Calculate $\theta_{DF} \leftarrow (\theta_{SK} - \theta_{FT})$, $\dot{\theta}_{DF} \leftarrow (\dot{\theta}_{SK} - \dot{\theta}_{FT})$
7       Fill data windows $\vec{W}_{\theta_{SK}}, \vec{W}_{\theta_{DF}}$ with angles of $\theta_{SK}$ and $\theta_{DF}$ measured in stance period of one cycle
8       **switch** $S_{pha}$ **do**
9         **case** *FootOff* **do**
10           **if** *EnterOnlyOncePerCycle* **then**
11             $I_{max} \leftarrow$ Find the index of max element in $\vec{W}_{\theta_{DF}}$
12             $\Delta\sigma_1 \leftarrow (\vec{W}_{\theta_{SK}}[I_{max}] - \vec{W}_{\theta_{SK}}[first]) \div 4 - \sigma_1(n-1)$   */\* $n-1$ means the value in last cycle \*/*
13             $\Delta\sigma_2 \leftarrow (\vec{W}_{\theta_{SK}}[end] - \vec{W}_{\theta_{SK}}[I_{max}]) \div 4 - \sigma_2(n-1)$   */\* see Equation (2)-(3)    \*/*
14             $\Delta\mu \leftarrow (\vec{W}_{\theta_{SK}}[I_{max}] - \mu[n-1])$
15             **if** *The values of $\Delta\sigma_1, \Delta\sigma_2, \Delta\mu$ are normal* **then**
16               Update parameters with a gain of 0.3 and generate $\vec{G}(n) = (\sigma_1(n), \sigma_2(n), \mu(n))$
17             **else**
18               $\vec{G}(n) \leftarrow \vec{G}(n-1)$
19             **end**
20             Clear $\vec{W}_{\theta_{SK}}, \vec{W}_{\theta_{DF}}$ data windows
21           **end**
22           **if** *Max foot angle has been detected* **then**
            */\* using a modified version of the previous Extremum Seeking Algorithm $(T\text{-}ASE, 19(3), 2022)$    \*/*
23             $S_{pha} \leftarrow$ FootContact
24           **end**
25         **end**
26         **case** *FootContact* **do**
27           **if** *Min foot angular velocity has been detected* **then**
28             $S_{pha} \leftarrow$ FootOff
29           **end**
30         **end**
31       **end**
32    **end**
33    Deinitialization and exit this thread

| **Child Thread 2:** ForceSensor - Get load cell measurement |
|---|

**Input:** flagToRunForceThread
**Output:** $F_{meas}$: load cell measurement [Unit: N]

1 Initialization
  /* create a child thread for each leg (one load cell), each thread performs the following codes    */
  /* while loop frequency: 1000 Hz    */
2 **while** *flagToRunForceThread* **do**
3     Check data availability
4     $F_{meas} \longleftarrow$ read and parse load cell measurement
5 **end**
6 Deinitialization and exit this thread

| **Child Thread 3:** RobotControl - Control the soft exoskeleton to generate desired forces on the user's body |
|---|

**Input:** flagToRunRobotThread
        $\theta_{SK}$: shank angle
        $\dot{\theta}_{SK}$: shank angular velocity
        $\theta_{DF}$: ankle dorsiflexion (DF) angle
        $\dot{\theta}_{DF}$: angle DF angular velocity
        $S_{pha}$: indicate foot contact or foot-off event
        $\vec{G}(n) = (\sigma_1, \sigma_2, \mu)$: estimated Gaussian parameters at $n$th gait cycle
        $F_{meas}$: load cell measurement

1 Initialization
2 Pretight actuation cable to confirm $L_1[t_0] + L_2[t_0]$  /* see Equation (6)    */
3 Release the cable for later transparent/silent walking  /* assistance isn't provided in the first five gaits    */
  /* while loop frequency: 1000 Hz    */
4 **while** *flagToRunRobotThread* **do**
  /* each side of legs alternately performs the following codes once during each while loop iteration    */
5     Ensure forces aren't beyond safe limits, otherwise reset all child thread flags
6     Get the length $L_{meas}$ and velocity $\dot{L}_{meas}$
7     **if** $S_{pha}$ *is FootOff* **then**
8        $L_{swing}[n] \longleftarrow$ Calculate the cable length for semi-slack state  /* see Equation (8)    */
9        $V_{swing,cmd} \longleftarrow$ Calculate velocity command based on PI control with damping  /* see Equation (9) */
10        $F_{swing,max}[n] \longleftarrow$ Monitor the maximum cable force throughout the swing period
11     **end**
12     **if** $S_{pha}$ *is FootContact* **then**
13        $F_{des}(\theta_{SK}) \longleftarrow$ Calculate desired force at the current shank angle $\theta_{SK}$  /* see Equation (1)    */
14        $V_{FB} \longleftarrow$ Calculate feedback component  /* see Equation (10)    */
15        Calculate $r \times (d\theta_{DF}/dt)$ and $(1/K_{all}) \times (dF_{des}(\theta)/dt)$  /* see Equations (12), (13)    */
16        $V_{FF} \longleftarrow$ Calculate feedforward component
17        $V_{stance,cmd} \longleftarrow$ Calculate velocity command  /* see Equation (11)    */
18        Monitor the maximum force and the corresponding shank angle throughout the stance period
19     **end**
20     Check the current motor position aren't or will be not beyond safe limits
21     Remain the actuation cable in fully slack state when in silent walking or when the foot remains static
22     Send $V_{cmd}$ ($V_{swing,cmd}$ or $V_{stance,cmd}$) to motor servo drivers
23 **end**
24 Zero the motor speed after the actuation cable being released to the initial position
25 Deinitialization and exit this thread

Article: A Shank Angle-Based Control System Enables Soft Exoskeleton to Assist Human Non-Steady Locomotion

**Data File S3. Matlab program for the experimental protocols.**

This file was uploaded on GitHub, and the link is [https://github.com/xw-TAN/SK_Exp_Protocol](https://github.com/xw-TAN/SK_Exp_Protocol).